\pdfoutput=1
\documentclass[11pt]{article}

\usepackage[final]{acl}
\usepackage{times}
\usepackage{latexsym}
\usepackage[T1]{fontenc}
\usepackage[utf8]{inputenc}
\usepackage{microtype}
\usepackage{inconsolata}
\usepackage{graphicx}
\usepackage{booktabs}
\usepackage{amsmath, amssymb, amsfonts}
\usepackage{algorithm}
\usepackage{algorithmic}
\usepackage{caption}
\usepackage{hyperref}
\usepackage{listings}
\usepackage{listingsutf8}
\newcommand\blfootnote[1]{%
  \begingroup
  \renewcommand\thefootnote{}\footnote{#1}%
  \addtocounter{footnote}{-1}%
  \endgroup
}

\title{Aryabhata: An exam-focused language model for JEE Math}

\author{Ritvik Rastogi \\
  PhysicsWallah \\
  \texttt{ritvik.rastogi@pw.live} \\\And
  Sachin Dharashivkar \\
  AthenaAgent \\
  \texttt{sachin@athenaagent.com} \\\And
  Sandeep Varma \\
  PhysicsWallah \\
  \texttt{sandeep.varma@pw.live}
  }

\begin{document}
\maketitle

\begin{abstract}
We present \textbf{Aryabhata 1.0}, a compact 7B parameter math reasoning model optimized for the Indian academic exam, the Joint Entrance Examination (JEE). Despite rapid progress in large language models (LLMs), current models often remain unsuitable for educational use. Aryabhata 1.0 is built by merging strong open-weight reasoning models, followed by supervised fine-tuning (SFT) with curriculum learning on verified chain-of-thought (CoT) traces curated through best-of-$n$ rejection sampling. To further boost performance, we apply reinforcement learning with verifiable rewards (RLVR) using A2C objective with group-relative advantage estimation alongwith novel exploration strategies such as \textit{Adaptive Group Resizing} and \textit{Temperature Scaling}. Evaluated on both in-distribution (JEE Main 2025) and out-of-distribution (MATH, GSM8K) benchmarks, Aryabhata outperforms existing models in accuracy and efficiency, while offering pedagogically useful step-by-step reasoning. We release Aryabhata as a foundation model to advance exam-centric, open-source small language models. \textbf{This marks our first open release for community feedback (\href{https://huggingface.co/PhysicsWallahAI/Aryabhata-1.0}{Aryabhata 1.0 on Hugging Face}); PW is actively training future models to further improve learning outcomes for students.} \blfootnote{Correspondence to \texttt{ritvik.rastogi@pw.live}}
\end{abstract}

\section{Introduction}

Large language models (LLMs) have shown remarkable progress in mathematical reasoning, but most existing systems fall short in supporting student learning in academic settings like India’s Joint Entrance Examination (JEE). These exams require not only accurate solutions but also transparent and precise reasoning that aids student understanding and long-term learning.

We observe three broad classes of models in this space:

\textbf{Non-reasoning models} Instruction-tuned models (e.g., GPT-4o) were largely inaccurate on rigorous math exams like JEE. These models failed to perform multi-step reasoning, often guessing answers or relying on shallow pattern matching.

\textbf{Early reasoning models} introduced long chain-of-thought (CoT) reasoning to improve accuracy, with examples including OpenAI o1 \citep{openai2024o1} and DeepSeek R1 \citep{deepseekai2025deepseekr1incentivizingreasoningcapability}. While these models were more accurate than non-reasoning baselines, they remained impractical in real-world educational settings. For instance, o1 \citep{openai2024o1} did not expose its reasoning trace and provided just a summary of them, while DeepSeek R1 \citep{deepseekai2025deepseekr1incentivizingreasoningcapability} produced long, nonlinear traces that made it difficult for students to follow the logic. Moreover, both models were relatively slow, generating lengthy explanations that consumed a significant amount of tokens and latency.

\textbf{Modern reasoning models} such as OpenAI o4-mini \citep{openai2025o3o4minirelease}, Gemini 2.5 \citep{comanici2025gemini25pushingfrontier}, and the updated version of DeepSeek R1 \citep{deepseekai2025deepseekr1incentivizingreasoningcapability} have improved further in raw accuracy and generation speed. However, pedagogical usability remains limited. For example, o4-mini \citep{openai2025o3o4minirelease} provides just a summary of its reasoning traces, while Gemini \citep{comanici2025gemini25pushingfrontier} and DeepSeek R1 \citep{deepseekai2025deepseekr1incentivizingreasoningcapability} still produce nonlinear, self-correcting reasoning paths that confuse learners rather than clarify concepts. (Samples are provided in Appendix \ref{app:examples}.)

In this work, we present \textbf{Aryabhata 1.0}, a compact and open 7B parameter model tailored for math reasoning in Indian competitive exams. Built via model merging and fine-tuned with domain-aligned data, Aryabhata combines accuracy, transparency, and efficiency, making it a viable foundation for educational AI applications.

\section{Related Work}

Current math LLMs built on open‑weight backbones have primarily leveraged Imitation Learning, Supervised Fine Tuning, and Reinforcement Learning to enhance chain-of-thought mathematical reasoning.

For instance \textbf{DeepSeekMath} \citep{shao2024deepseekmathpushinglimitsmathematical}, introduced in early 2024, advanced the capabilities of open weight models by pre-training on hundreds of billions of math‑focused tokens and pioneering Group Relative Policy Optimization (GRPO).

\textbf{Qwen‑2.5‑Math‑7B} \citep{yang2024qwen25mathtechnicalreportmathematical} is a math‑specialized 7B instruction‑tuned model that supports chain‑of‑thought (CoT) and tool‑integrated reasoning (TIR) across both English and Chinese problem sets.

NVIDIA's \textbf{AceMath‑7B‑Instruct} \citep{liu2025acemathadvancingfrontiermath}, derived from Qwen, advances its performance further through a multi-stage SFT training pipeline designed to improve both mathematical and reasoning accuracy  on multiple benchmarks and edging close to much larger models at 72B scale.

Meanwhile, \textbf{DeepSeek‑R1} \citep{deepseekai2025deepseekr1incentivizingreasoningcapability} introduced a pure RL-based reasoning model trained with GRPO-style verifiable rewards, achieving impressive results. Its distilled variants (\textbf{DeepSeek‑R1‑Distill‑Qwen‑7B} \citep{deepseekai2025deepseekr1incentivizingreasoningcapability}) inherit reasoning performance via long CoT.

The \textbf{AceReason‑Nemotron‑7B} \citep{liu2025acemathadvancingfrontiermath} demonstrates that large-scale reinforcement learning can significantly enhance the reasoning capabilities of strong small- and mid-sized models by first training on math-only prompts, then on code-only prompts.

The \textbf{AceReason‑Nemotron‑1.1‑7B} \citep{liu2025acereasonnemotron11advancingmath} synergizes SFT and RL fine‑tuning by employing a stage-wise RL approach on math-only and code-only prompts.

Our approach builds on these lines by merging models for hybrid capabilities (symbolic fluency + coherent CoT), followed by rejection‑sampled SFT and RL with verifiable rewards, preserving both performance and efficiency in a compact model.

\section{Methodology}
The overall process can be categorized in the following four stages:

\subsection{Model Merging}
The development of LLMs has seen a transition from System 1 (quick thinking) to System 2 (deliberate, methodical) reasoning, each with distinct advantages \citep{wu2025unlockingefficientlongtoshortllm}. While System 1 models excel at producing fluent answers with low latency, they often lack the depth required for complex reasoning. In contrast, System 2 models are capable of iterative self-correction and structured reasoning, but suffer from inefficiencies due to verbose or redundant CoT traces.

To address this challenge, Kimi k1.5 \citep{kimiteam2025kimik15scalingreinforcement} introduced the concept of merging reasoning and non-reasoning models, which was further explored by \citet{wu2025unlockingefficientlongtoshortllm}. Building on this intuition, we carefully selected three distinct LLMs, each sharing the same base architecture.
\begin{itemize}
    \item Qwen2.5-Math-7B-Instruct \citep{yang2024qwen25mathtechnicalreportmathematical}, a strong open source mathematical LLM providing solid baseline capabilities and fundamental math fluency.
    \item AceMath-7B-Instruct \citep{liu2025acemathadvancingfrontiermath}  a version of Qwen 2.5 Math that was further fine-tuned by NVIDIA, significantly enhancing its accuracy on mathematical benchmarks.
    \item DeepSeek-R1-Distill-Qwen-7B \citep{deepseekai2025deepseekr1incentivizingreasoningcapability}, a long-form reasoning model derived by fine-tuning Qwen 2.5 Math on reasoning traces distilled from DeepSeek R1 \citep{deepseekai2025deepseekr1incentivizingreasoningcapability}.
\end{itemize}

We apply linear merging \citep{wortsman2022modelsoupsaveragingweights} to combine the models using the MergeKit \citep{goddard-etal-2024-arcees} framework.

Let $\theta_1$, $\theta_2$, $\theta_3$ be the parameters of Qwen, Ace, and DeepSeek, respectively. We compute:

\[
\theta_{\text{merged}} = \alpha \theta_1 + \beta \theta_2 + \gamma \theta_3,\quad \text{where } \alpha + \beta + \gamma = 1
\]

We select weights $\alpha, \beta, \gamma$ empirically based on the held-out math reasoning tasks. Final weights favor quickly addressing simpler problems while also performing methodical, multi-step analysis for more complex mathematical challenges.

\begin{table}[h]
\centering
\begin{tabular}{lc}
\hline
\textbf{Topic} & \textbf{\%age} \\
\hline
Application of Derivatives & 4.50\% \\
Application of Integrals & 2.27\% \\
Binomial Theorem & 2.37\% \\
Circles & 2.85\% \\
Complex Numbers \& \\ \hspace{2cm}Quadratic Equations & 6.00\% \\
Conic Section & 7.55\% \\
Continuity and Differentiability & 2.71\% \\
Definite Integration & 2.45\% \\
Determinants & 3.04\% \\
Differential Equations & 3.77\% \\
Indefinite Integration & 3.26\% \\
Inverse Trigonometric Functions & 5.31\% \\
Limits and Derivatives & 3.88\% \\
Matrices & 2.46\% \\
Permutations and Combinations & 4.23\% \\
Probability & 5.69\% \\
Quadratic Equations & 4.45\% \\
Relations and Functions & 2.24\% \\
Sequence and Series & 2.75\% \\
Sets & 1.04\% \\
Statistics & 1.89\% \\
Straight Lines & 2.31\% \\
Three Dimensional Geometry & 3.92\% \\
Trigonometric Functions & 4.51\% \\
Vector Algebra & 2.89\% \\
Miscellaneous & 11.65\% \\
\hline
\end{tabular}
\caption{Topic-wise Question Distribution}
\label{tab:topic_distribution}
\end{table}

\subsection{Data Curation}

High-quality, domain-aligned data is essential for training effective mathematical reasoning models. To this end, we relied on a proprietary corpus curated by the subject matter experts and educators at PhysicsWallah, ensuring close alignment with the Indian Joint Entrance Examination (JEE) standards. This dataset represents years of academic effort and is considered the core intellectual property of PhysicsWallah. As such, we do not publicly release the training data.

We parsed approximately 250,000 raw questions from internal databases. To ensure syntactic coherence and semantic relevance, we applied the following filtering steps:
\begin{itemize}
    \item Removed diagram-based questions, which require multimodal reasoning not supported by current text-only models.
    \item Filtered out non-English or poorly formatted questions.
    \item Stripped all answer options from the remaining questions to frame the task as open-ended generation rather than classification. This design choice was also explored by \citet{chandak2025answermatchingoutperformsmultiple}
    \item Since we stripped options from the questions, we removed the questions which relied on options to be answered such as "which of the following is true"
\end{itemize}

To standardize and clean raw question-answer pairs, we designed a structured prompt (see Appendix~\ref{app:filter}) that extracts the core question, normalizes the answer format, identifies dependencies and detects the question language, using OpenAI o4-mini.

This process resulted in a clean dataset of around 130,000 questions suitable for the generation of further chain of thought. The topic-wise distribution of questions is outlined in Table \ref{tab:topic_distribution}.

\begin{table*}[t]
\centering
\begin{tabular}{cccc}
\toprule
\textbf{Correct CoTs} & \textbf{\# Questions} & \textbf{Total CoTs} & \textbf{Usage} \\
\midrule
0 & 31,470 & 0 & Used in RLVR only \\
1 & 9,647 & 9,647 & SFT \\
2 & 9,066 & 18,132 & SFT \\
3 & 12,643 & 37,929 & SFT \\
4 & 67,247 & 268,988 & 10\% sampled for SFT \\
\bottomrule
\end{tabular}
\caption{Chain-of-Thought generation outcomes from best-of-4 sampling.}
\label{tb: rejection_sampling}
\end{table*}

\subsection{Supervised Fine-Tuning with Rejection Sampling}

To generate high-quality chain-of-thought (CoT) supervision, we employed best-of-4 rejection sampling using the merged model. For each curated question $x$, we sampled four CoT responses $\{y_1, y_2, y_3, y_4\}$, and selected only those completions whose final answer matched the known correct answer i.e. $\text{GT}(x)$, using Algorithm~\ref{alg: answer matching procedure}. This filtering process ensures logical correctness and minimizes noisy supervision signals.

We then grouped the questions based on how many of the four generations lead to the correct answers and selected samples for curriculum-style supervised fine-tuning \citep{bengio2009curriculumlearning}, i.e., beginning the training with easier samples (e.g., 4/4 correct) and gradually introducing harder examples (e.g., 3/4, 2/4, 1/4 correct). This curriculum-based training stabilizes early learning and improves generalization on harder problems.

Let $\mathcal{D}_{\text{sft}} = \{(x^{(i)}, y^{(i)})\}_{i=1}^{N}$ denote the dataset of input questions and their corresponding verified CoT completions. The supervised fine-tuning objective minimizes the standard next-token prediction loss:

\begin{equation}
\mathcal{L}_{\text{SFT}} = -\sum_{(x, y) \in \mathcal{D}_{\text{sft}}} \sum_{t=1}^{T} \log (p_{\theta}(y_t \mid x, y_{<t}))
\end{equation}

where $y_t$ is the $t^{\text{th}}$ token of the target CoT sequence, and $p_{\theta}$ is the model's probability distribution parameterized by $\theta$.

In total, we obtained approximately 350,000 verified CoTs across around 100,000 questions, which were sampled to serve as the core training corpus for SFT, as detailed in Table~\ref{tb: rejection_sampling}. The 0/4 cases were retained for downstream reinforcement learning with verifiable rewards (RLVR) to further improve coverage and robustness in challenging problem spaces.

We used Parameter Efficient Finetuning, particulary Low-Rank Adaptation \citep{hu2021loralowrankadaptationlarge} during SFT using peft \citep{peft} library, the training parameters are mentioned in Appendix \ref{app:sft_hp}.

\subsection{Reinforcement Learning with Verifiable Rewards}

We extend Reinforcement Learning with Verifiable Rewards (RLVR) \citep{lambert2025tulu3pushingfrontiers} by incorporating group-based advantage estimation \citep{shao2024deepseekmathpushinglimitsmathematical} within an Advantage Actor-Critic (A2C) framework \citep{mnih2016asynchronousmethodsdeepreinforcement} .

\subsubsection{Group-Relative Policy Optimization}

Our approach optimizes the following A2C objective with group-relative advantage estimation:

\begin{align*}
\hspace{0.5cm} J^{A2C}(\theta) = & \\ & \hspace{-1cm}  \mathbb{E}{(\alpha_i) \sim \pi\theta} \left[ \frac{1}{G} \sum_{i=1}^{G} \frac{1}{|\alpha_i|} \log \pi_\theta(\alpha_i) \cdot \tilde{A}_i \right]
\end{align*}

We optimize the A2C objective over G sampled response sequences $\alpha_i$, applying length-normalized gradients weighted by sequence-level advantages $\tilde{A}_i$ computed through group-relative advantage estimation.

\textbf{Binary Reward Structure}: We employ a simple binary reward that provides unambiguous feedback for mathematical reasoning:

$$R_i = \begin{cases}
1 & \text{if the final answer is correct} \\
0 & \text{if the final answer is incorrect}
\end{cases}$$

\textbf{Group Advantage Estimation} The advantage function is computed using group-relative normalization:

$\hat{A}_{i,t} = \frac{R_i - \bar{R}_{\text{group}}}{\sigma_{\text{group}}}$

where $\bar{R}_{\text{group}}$ is the mean reward across all solutions in the group and $\sigma_{\text{group}}$ is the standard deviation.

\textbf{Key Benefits}: This group-relative baseline offers several advantages:
\begin{itemize}
    \item \textbf{Reduced variance}: Group comparison stabilizes gradient estimates
    \item \textbf{Simplified training}: Eliminates need for KL divergence constraints or probability ratio clipping
    \item \textbf{Natural compatibility}: Works seamlessly with binary rewards, common in mathematical reasoning tasks    
\end{itemize}

\subsubsection{Exploration Strategies}
\textbf{Adaptive Group Sizing}: Unlike fixed group sizes in standard GRPO implementations (\citet{vonwerra2022trl}, \citet{sheng2024hybridflow}, \citet{unsloth}), we dynamically adjust group size based on problem difficulty. Starting with a group size of 8 for simpler problems, we scale up to a group size of 64 for harder ones.

The dynamic group size follows:
$$G_d = 8 \times 2^k$$
where $k \in \{0, 1, 2, 3\}$ is determined by the group average reward $\bar{R}_{\text{group}}$. When performance drops below preset thresholds, we increase $k$, scaling groups as: $8 \rightarrow 16 \rightarrow 32 \rightarrow 64$.

This adaptive scaling improves sampling diversity and advantage estimation stability for challenging problems while efficiently allocating computational resources.

\textbf{Progressive Temperature Scaling}: We continuously increase the sampling temperature from 0.6 to 1.0 throughout training, this was explored in contemporary works like POLARIS \citep{Polaris2025}. This progressive scaling balances exploitation and exploration:

\begin{itemize}
    \item \textbf{Initial phase}: Low temperature (0.6) promotes training stability through conservative sampling
    \item \textbf{Progressive increase}: Temperature gradually rises, encouraging more diverse solution exploration
    \item \textbf{Final phase}: Temperature reaches 1.0, enabling much more exploration of the action space compared to lower temperatures.
\end{itemize}

\textbf{Curriculum-Based Sampling}: 
We filter training samples to focus on an optimal difficulty range, removing both trivial and intractable problems:

\begin{itemize}
    \item \textbf{Too easy}: Provide minimal learning signal due to high success rates
    \item \textbf{Too hard}: Introduce noise through consistently low performance

\end{itemize}

Our filtering uses a difficulty assessment function $f_{\text{difficulty}}(x)$ based on model success rates:

$$\mathcal{D}_t^{\text{filtered}} = \{x \in \mathcal{D}_t : \alpha_{\text{min}} \leq f_{\text{difficulty}}(x) \leq \alpha_{\text{max}}\}$$

This curriculum approach concentrates computational resources on problems that maximize learning progress.

\subsubsection{Training Configuration and Hyperparameters}

Our reinforcement learning implementation employs carefully tuned hyperparameters optimized for mathematical reasoning tasks while maintaining computational efficiency within hardware constraints. The training configuration incorporates modern optimization techniques and memory-efficient strategies to enable stable convergence at scale.

\textbf{Optimization Configuration}: We utilize the Adam optimizer \citep{kingma2017adammethodstochasticoptimization} with a conservative learning rate of $1 \times 10^{-6}$ to ensure stable policy gradient updates throughout the training process.

\textbf{Memory and Precision Management}: Training is conducted using bfloat16 (BF16) mixed precision arithmetic, which provides substantial memory savings while maintaining numerical stability for gradient computations. Gradient checkpointing is employed to further reduce memory consumption during backpropagation, enabling training of larger models within available GPU memory constraints.

\textbf{Sequence and Batch Configuration}: The model operates within a maximum context length of 4,096 tokens, providing sufficient capacity for complex multi-step mathematical reasoning while maintaining computational tractability. 

\section{Evaluation}
We evaluated Aryabhata 1.0 across both in-distribution and out-of-distribution math benchmarks to assess its accuracy and efficiency in solving problems at scale.

We evaluate model-generated solutions using the \texttt{pass@1} accuracy. The solutions are generated using greedy decoding (temperature = 0). To determine whether a predicted answer matches the ground-truth answer for a question, we follow the pipeline described in the Algorithm~\ref{alg: answer matching procedure}.

\begin{algorithm}
\caption{Answer Matching Procedure}
\label{alg: answer matching procedure}
\begin{algorithmic}[1]
\STATE \textbf{Input:} Predicted answer $a_p$, Ground-truth answer $a_g$, Options (if any)
\STATE \textbf{Output:} Match status (True / False)

\IF{$a_p = a_g$ or sympy\_latex\_match($a_p$, $a_g$)}
    \RETURN True
\ENDIF
\IF{option/identifier from $a_p$ == option/identifier from $a_g$}
    \RETURN True
\ENDIF
\STATE Query LLM judge with $a_p$, $a_g$, and options (if any)
\IF{LLM returns YES}
    \RETURN True
\ELSE
    \RETURN False
\ENDIF
\end{algorithmic}
\end{algorithm}

Depending on whether the question is Multiple Choice Question or a Numerical Answer Type, we use different prompts to query the judge model (GPT-4o-mini). The prompts are provided in Table \ref{tb: answer_matching}.

\subsection{In-Distribution Evaluation: JEE Main 2025}

\begin{figure*}[htbp]
    \centering
    \includegraphics[width=\textwidth]{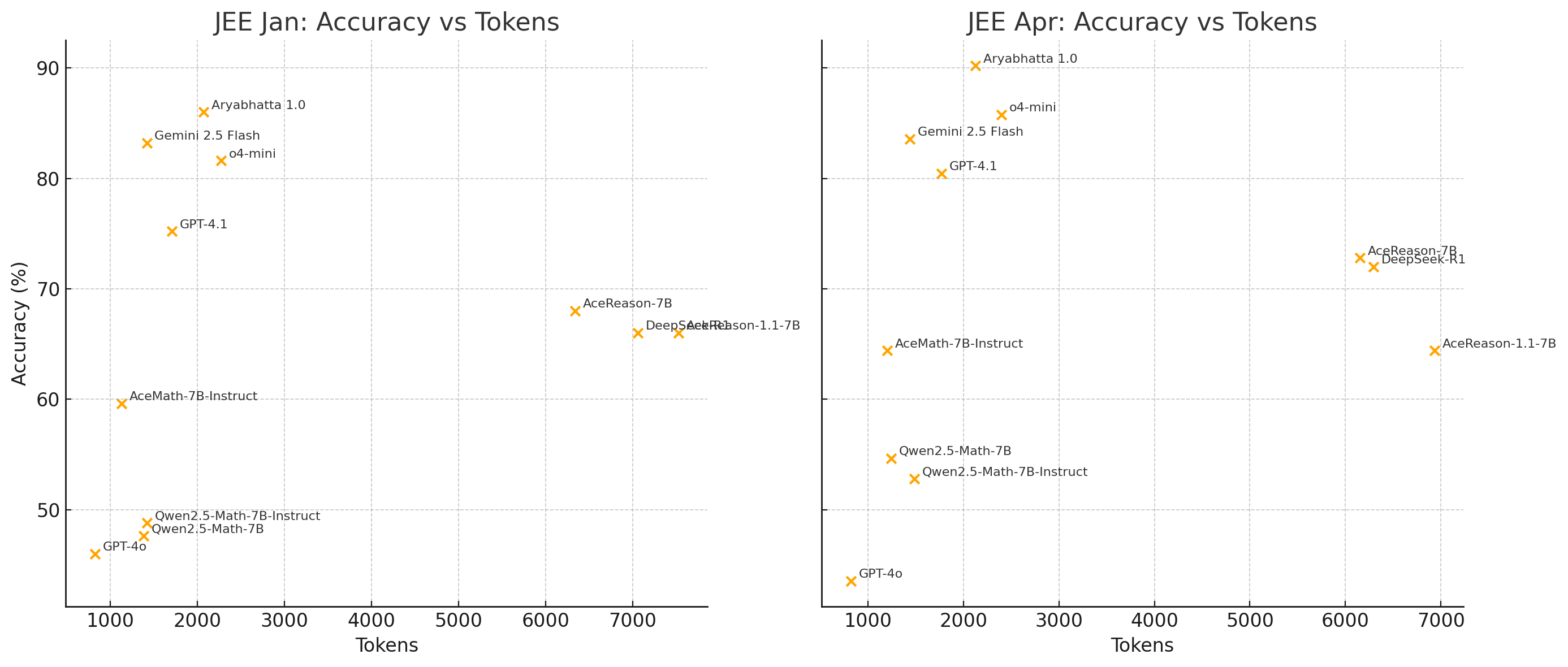} % or .png
    \caption{Scatter plots showing Accuracy vs. Tokens for JEE Jan and JEE Apr.}
    \label{fig:accuracy-tokens}
\end{figure*}

\begin{table*}[h]
\centering
\begin{tabular}{lcc}
\hline
\textbf{Model} & \textbf{MATH 500} & \textbf{GSM8K} \\
\hline
Aryabhatta 1.0 & 83.6 & 94.8 \\
Qwen/Qwen2.5-Math-7B-Instruct & 66.0 & 94.7 \\
nvidia/AceMath-7B-Instruct & 80.6 & 93.4  \\
GPT-4o & 69.2 & 94.6 \\
deepseek-ai/DeepSeek-R1-Distill-Qwen-7B & 85.2 & 69.7 \\
nvidia/AceReason-Nemotron-7B & 84.2 & 76.5 \\
nvidia/AceReason-Nemotron-1.1-7B & 85.4 & 93.1 \\
GPT-4.1 & 86.6 & 94.0 \\
o4-mini & 94.8 & 90.1 \\
Gemini 2.5 Flash & 93.6 & 85.1 \\
\hline
\end{tabular}
\caption{Performance comparison on MATH 500 and GSM8K benchmarks}
\label{tab:model_comparison}
\end{table*}

To measure performance in familiar distribution settings, we evaluate Aryabhata on the JEE Main 2025 exam. The January session contains 250 questions (10 papers with 25 questions each), while the April session comprises 225 questions (9 papers with 25 questions each), all sourced from official exam papers.

Figure \ref{fig:accuracy-tokens} shows that Aryabhata 1.0 achieves an accuracy of \textbf{86.0\%} on the January session and \textbf{90.2\%} on the April session, while maintaining token efficiency with an average of approximately \textasciitilde{}2K tokens per response.

Compared to both open-weight and proprietary models, Aryabhata outperforms all baselines in accuracy while remaining competitive in inference cost.

\subsection{Out-of-Distribution Evaluation}

To evaluate generalization beyond the fine-tuning distribution, we benchmark Aryabhata 1.0 on the following datasets:

\begin{itemize}
    \item \textbf{MATH 500}: A curated benchmark of 500 competition-style problems drawn from the larger MATH dataset originally introduced by \citet{hendrycks2021measuringmathematicalproblemsolving}.
    \item \textbf{GSM8K} \citep{cobbe2021trainingverifierssolvemath}: A widely used benchmark of grade school math word problems.
\end{itemize}

Table \ref{tab:model_comparison} shows that Aryabhata demonstrates \textbf{competitive generalization} to unseen tasks of comparable difficulty, outperforming its base models on both MATH and GSM8K. 

\section*{Conclusion and Future Work}

In this work, we introduced \textbf{Aryabhata 1.0}, a compact open source model with 7B parameters for mathematical reasoning, specifically designed for the Indian competitive exam ecosystem. By merging diverse mathematical LLMs and fine-tuning on carefully curated and verified domain-specific data, Aryabhata achieves state-of-the-art performance on in-distribution benchmarks such as JEE Main, while demonstrating competitive generalization to out-of-distribution tasks like MATH and GSM8K.

Looking ahead, we plan to:
Expand coverage to Physics and Chemistry, building similar reasoning capabilities in other STEM domains. Scale to the full syllabus across Foundation, JEE (Main \& Advanced), and NEET, enabling end-to-end subject-level assistance. Develop a family of exam-centric, open source small language models (SLMs) that are compact, efficient, and aligned to Indian education standards.

We believe that this direction will empower millions of students with accessible and curriculum-aligned AI tools that complement classroom learning and personalized preparation.

\section*{Acknowledgments}

We thank \textbf{Tejas Chaudhari} and \textbf{Vishal Singh} for their efforts in creating evaluation datasets and writing model evaluation scripts. \textbf{Shubham Choudhari} contributed to building the training pipelines and running reinforcement learning experiments. \textbf{Archit Singhai} and \textbf{Chinmay Karkar} were instrumental in exploring and benchmarking existing RL libraries.

\bibliography{main}

\begin{thebibliography}{25}
\providecommand{\natexlab}[1]{#1}

\bibitem[{An et~al.(2025)An, Xie, Li, Li, Zhang, Gong, Zhong, Xu, Qiu, Wang, and Kong}]{Polaris2025}
Chenxin An, Zhihui Xie, Xiaonan Li, Lei Li, Jun Zhang, Shansan Gong, Ming Zhong, Jingjing Xu, Xipeng Qiu, Mingxuan Wang, and Lingpeng Kong. 2025.
\newblock \href {https://hkunlp.github.io/blog/2025/Polaris} {Polaris: A post-training recipe for scaling reinforcement learning on advanced reasoning models}.

\bibitem[{Bengio et~al.(2009)Bengio, Louradour, Collobert, and Weston}]{bengio2009curriculumlearning}
Yoshua Bengio, Jérôme Louradour, Ronan Collobert, and Jason Weston. 2009.
\newblock \href {https://doi.org/10.1145/1553374.1553380} {Curriculum learning}.
\newblock In \emph{Proceedings of the 26th International Conference on Machine Learning (ICML)}, volume 382 of \emph{ACM International Conference Proceeding Series}, pages 41--48. ACM.

\bibitem[{Chandak et~al.(2025)Chandak, Goel, Prabhu, Hardt, and Geiping}]{chandak2025answermatchingoutperformsmultiple}
Nikhil Chandak, Shashwat Goel, Ameya Prabhu, Moritz Hardt, and Jonas Geiping. 2025.
\newblock \href {https://arxiv.org/abs/2507.02856} {Answer matching outperforms multiple choice for language model evaluation}.
\newblock \emph{Preprint}, arXiv:2507.02856.

\bibitem[{Cobbe et~al.(2021)Cobbe, Kosaraju, Bavarian, Chen, Jun, Kaiser, Plappert, Tworek, Hilton, Nakano, Hesse, and Schulman}]{cobbe2021trainingverifierssolvemath}
Karl Cobbe, Vineet Kosaraju, Mohammad Bavarian, Mark Chen, Heewoo Jun, Lukasz Kaiser, Matthias Plappert, Jerry Tworek, Jacob Hilton, Reiichiro Nakano, Christopher Hesse, and John Schulman. 2021.
\newblock \href {https://arxiv.org/abs/2110.14168} {Training verifiers to solve math word problems}.
\newblock \emph{Preprint}, arXiv:2110.14168.

\bibitem[{Comanici et~al.(2025)Comanici, Bieber, Schaekermann, Pasupat, Sachdeva, Dhillon, Blistein, Ram, Zhang, Rosen, Marris, Petulla, Gaffney, Aharoni, Lintz, Pais, Jacobsson, Szpektor, Jiang, Haridasan, Omran, Saunshi, Bahri, Mishra, Chu, Boyd, Hekman, Parisi, Zhang, Kawintiranon, Bedrax-Weiss, Wang, Xu, Purkiss, Mendlovic, Deutel, Nguyen, Langley, Korn, Rossazza, Ramé, Waghmare, Miller, Byrd, Sheshan, Bhardwaj, Janus, Rissa, Horgan, Silver, Wahid, Brin, Raimond, Kloboves, Wang, Gundavarapu, Shumailov, Wang, Pajarskas, Heyward, Nikoltchev, Kula, Zhou, Garrett, Kafle, Arik, Goel, Yang, Park, Kojima, Mahmoudieh, Kavukcuoglu, Chen, Fritz, Bulyenov, Roy, Paparas, Shemtov, Chen, Strudel, Reitter, Roy, Vlasov, Ryu, Leichner, Yang, Mariet, Vnukov, Sohn, Stuart, Liang, Chen, Rawlani, Koh, Co-Reyes, Lai, Banzal, Vytiniotis, Mei, Cai, Badawi, Fry, Hartman, Zheng, Jia, Keeling, Louis, Chen, Robles, Hung, Zhou, Saxena, Goenka, Ma, Fisher, Taege, Graves, Steiner, Li, Nguyen, Sukthankar, Stanton, Eslami, Shen, Akin,
  Guseynov, Zhou, Alayrac, Joulin, Farkash, Thapliyal, Roller, Shazeer, Davchev, Koo, Forbes-Pollard, Audhkhasi, Farquhar, Gilady, Song, Aslanides, Mendolicchio, Parrish, Blitzer, Gupta, Ju, Yang, Datta, Tacchetti, Mehta, Dibb, Gupta, Piccinini, Hadsell, Rajayogam, Jiang, Griffin, Sundberg, Hayes, Frolov, Xie, Zhang, Dasgupta, Kalra, Shani, Macherey, Huang, MacDermed, Duddu, Zacchello, Yang, Lo, Hui, Kastelic, Gasaway, Tan, Yue, Barrio, Wieting, Yang, Nystrom, Demmessie, Levskaya, Viola, Tekur, Billock, Necula, Joshi, Schaeffer, Lokhande, Sorokin, Shenoy, Chen, Collier, Li, Bos, Wichers, Lee, Pouget, Thangaraj, Axiotis, Crone, Sterneck, Chinaev, Krakovna, Ferludin, Gemp, Winkler, Goldberg, Korotkov, Xiao, Mehrotra, Mariserla, Piratla, Thurk, Pham, Ma, Senges, Kumar, Meyer, Talius, Pierse, Sandhu, Toma, Lin, Nath, Stone, Sadigh, Gupta, Guez, Singh, Thomas, Duerig, Gong, Tanburn, Zhang, Dao, Hammad, Xie, Rijhwani, Murdoch, Kim, Thompson, Cheng, Sohn, Sprechmann, Xu, Tadepalli, Young, Zhang, Srinivasan,
  Aperghis, Ayyar, Fitoussi, Burnell, Madras, Dusenberry, Xiong, Oguntebi, Albrecht, Bornschein, Mitrović, Dimarco, Shamanna, Shah, Sezener, Upadhyay, Lacey, Schiff, Baur, Ganapathy, Schnider, Wirth, Schenck, Simanovsky, Tan, Fränken, Duan, Mankalale, Dhawan, Sequeira, Wei, Goel, Unlu, Zhu, Sun, Balashankar, Shuster, Umekar, Alnahlawi, van~den Oord, Chen, Zhai, Dai, Lee, Doi, Zilka, Vallu, Shrivastava, Lee, Husain, Zhuang, Cohen-Addad, Barber, Atwood, Sadovsky, Wellens, Hand, Rajendran, Turker, Carey, Xu, Soltau, Li, Song, Li, Kemaev, Brown, Burns, Patraucean, Stanczyk, Aravamudhan, Blondel, Noga, Blanco, Song, Isard, Sharma, Hayes, Badawy, Lamp, Laish, Kozlova, Chan, Singla, Sunkara, Upadhyay, Liu, Bai, Wilkiewicz, Zlocha, Liu, Li, Li, Barak, Raboshchuk, Choi, Liu, Jue, Sharma, Marzoca, Busa-Fekete, Korsun, Elisseeff, Shen, Carthy, Lamerigts, Hosseini, Lin, Chen, Yang, Chauhan, Omernick, Jia, Zainullina, Hassabis, Vainstein, Amid, Zhou, Votel, Vértes, Li, Zhou, Lazaridou, McMahan, Narayanan, Soyer, Basu,
  Lee, Perozzi, Cao, Berrada, Arya, Chen, Katrina, Xu, Lochbrunner, Hofer, Sharifzadeh, Wu, Goldman, Awasthi, Wang, Wu, Sha, Zhang, Mikuła, Graziano, Mcloughlin, Giannoumis, Namiki, Malik, Radebaugh, Hall, Leal-Cavazos, Chen, Sindhwani, Kao, Greene, Griffith, Welty, Montgomery, Yoshino, Yuan, Goodman, Michaely, Lee, Sawhney, Chen, Zheng, Shum, Savinov, Pot, Pak, Zadimoghaddam, Bhatnagar, Lewenberg, Kutzman, Liu, Katzen, Selier, Djolonga, Lepikhin, Xu, Liang, Tan, Schillings, Ersoy, Blois, Bandemer, Singh, Lebedev, Joshi, Brown, Palmer, Pathak, Jalan, Zubach, Lall, Parker, Gunjan, Rogulenko, Sanghai, Leng, Egyed, Li, Ivanova, Andriopoulos, Xie, Rosenfeld, Wright, Sharma, Geng, Wang, Kwei, Pan, Zhang, Wang, Liu, Yeung, Cole, Rosenberg, Yang, Chen, Polovets, Nair, Saxena, Smith, yiin Chang, Mahendru, Grant, Iyer, Cai, McGiffin, Shen, Walton, Girgis, Woodman, Ke, Kwong, Rouillard, Rao, Li, Xu, Prost, Zou, Ji, Magni, Liechty, Calian, Ramachandran, Krivokon, Huang, Chen, Hauth, Ilić, Xi, Lim, Ion, Moradi,
  Toksoz-Exley, Bullard, Allamanis, Yang, Wang, Hong, Gergely, Li, Mittal, Kovalev, Ungureanu, Labanowski, Wassenberg, Lacasse, Cideron, Dević, Marsden, Nguyen, Fink, Zhong, Kiyono, Ivanov, Ma, Bain, Yalasangi, She, Petrushkina, Lunayach, Bromberg, Hodkinson, Meshram, Vlasic, Kyker, Xu, Stanway, Yang, Zhao, Tung, Odoom, Fujii, Gilmer, Kim, Halim, Le, Bohnet, El-Sayed, Neyshabur, Reynolds, Reich, Xu, Moreira, Sharma, Liu, Hosseini, Raisinghani, Su, Lao, Formoso, Gelmi, Gueta, Dey, Gribovskaya, Ćevid, Mudgal, Bingham, Wang, Kumar, Cullum, Han, Bousmalis, Cedillo, Chu, Magay, Michel, Hlavnova, Calandriello, Ariafar, Yao, Sehwag, Vezer, Lago, Zhu, Rubenstein, Porter, Baddepudi, Riva, Istin, Yeh, Li, Howard, Jha, Chen, de~Liedekerke, Ahmed, Rodriguez, Bhatia, Wang, Elqursh, Klinghoffer, Chen, Kohli, I, Zhang, Nado, Chen, Chen, Zhang, Singh, Hillier, Lebron, Tao, Liu, Dulac-Arnold, Zhang, Narayan, Liu, Firat, Bhowmick, Liu, Zhang, Zhang, Rotival, Howard, Sinha, Grushetsky, Beyret, Gopalakrishnan, Zhao, He,
  Payrits, Nabulsi, Zhang, Chen, Lee, Fallen, Gollapudi, Zhou, Pavetić, Köppe, Huang, Pasumarthi, Fernando, Fischer, Ćurko, Gao, Svensson, Stone, Qureshi, Sinha, Kulshreshtha, Matysiak, Mao, Saroufim, Faust, Duan, Fidel, Katircioglu, Kaufman, Shah, Kong, Bapna, Weisz, Dunleavy, Dutta, Liu, Chaabouni, Parada, Wu, Belias, Bissacco, Fort, Xiao, Huot, Knutsen, Blau, Li, Prendki, Love, Chow, Charoenpanit, Shimokawa, Coriou, Gregor, Izo, Akula, Pinto, Hahn, Paulus, Guo, Sharma, Hsieh, Chukwuka, Hashimoto, Rauschmayr, Wu, Angermueller, Wang, Gerlach, Pliskin, Mirylenka, Ma, Baugher, Gale, Bijwadia, Rakićević, Wood, Park, Chang, Seal, Tar, Krasowiak, Song, Stephanov, Wang, Maggioni, Lin, Wu, Paul, Jiang, Agrawal, Piot, Feng, Kim, Doshi, Lai, Chuqiao, Xu, Vikram, Chelba, Krause, Zhuang, Rae, Denk, Collister, Weerts, Luo, Lu, Garnes, Gupta, Spitz, Hassidim, Liang, Shafran, Humphreys, Vassigh, Wallis, Shejwalkar, Perez-Nieves, Hornung, Tan, Westberg, Ly, Zhang, Farris, Park, Kosik, Cankara, Maksai, Xu, Cassirer,
  Caelles, Abdolmaleki, Chiang, Fabrikant, Shetty, He, Giménez, Hashemi, Panthaplackel, Kulizhskaya, Deshmukh, Pighin, Alazard, Jindal, Noury, S, Qin, Dotiwalla, Spencer, Babaeizadeh, Chen, Mehta, Lees, Leach, Koanantakool, Akolzin, Comanescu, Ahn, Svyatkovskiy, Mustafa, D'Ambrosio, Garlapati, Lamblin, Agarwal, Song, Sessa, Coquinot, Maggs, Masoom, Pitta, Wang, Morris-Suzuki, Porter, Jia, Dudek, R, Paduraru, Ansell, Bolukbasi, Lu, Ganeshan, Wang, Griffiths, Benenson, He, Swirhun, Papamakarios, Chawla, Sengupta, Wang, Milutinovic, Mordatch, Jia, Smith, Ng, Nigam, Young, Vušak, Hechtman, Goenka, Zipori, Ayoub, Popat, Acharya, Yu, Bloxwich, Song, Roit, Li, Boag, Nayakanti, Chandra, Ding, Mehta, Hope, Zhang, Shtacher, Badola, Nakashima, Sozanschi, Comşa, Žužul, Caveness, Odell, Watson, de~Cesare, Lippe, Lockhart, Verma, Chen, Sun, Zhuo, Shah, Gupta, Muzio, Niu, Zait, Singh, Gaba, Ye, Ramachandran, Saleh, Popa, Dubey, Liu, Javanmardi, Epstein, Hemsley, Green, Ranka, Cohen, Fu, Ghemawat, Borovik, Martens,
  Chen, Shyam, Pinto, Yang, Ţifrea, Du, Gong, Agarwal, Kim, Frank, Shah, Song, Deng, Mikhalap, Chatziprimou, Chung, Creswell, Zhang, Jun, Lebsack, Truong, Andačić, Yona, Fornoni, Rong, Toropov, Soudagar, Audibert, Zaiem, Abbas, Rusu, Potluri, Weng, Kementsietsidis, Tsitsulin, Peng, Ha, Jain, Latkar, Ivanov, McLean, GP, Venkataraman, Liu, Krishnan, D'sa, Yogev, Collins, Lee, Ho, Doersch, Yona, Gao, Ferreira, Ozturel, Muckenhirn, Zheng, Balasubramaniam, Bansal, van~den Driessche, Eiger, Haykal, Misra, Goyal, Martins, Leung, Valfridsson, Flynn, Bishop, Pang, Halpern, Yu, Moore, Yuvein, Zhu, Thiagarajan, Drori, Xiao, Dery, Jagerman, Lu, Ge, Aggarwal, Khare, Tran, Elyada, Alet, Rubin, Chou, Tian, Bai, Chan, Lew, Misiunas, Bilal, Ray, Raghuram, Castro-Ros, Carpenter, Zheng, Kilgore, Broder, Xue, Kallakuri, Dua, Yuen, Chien, Schultz, Agrawal, Tsarfaty, Hu, Kannan, Marcus, Kothari, Sun, Horn, Bošnjak, Naeem, Hirsch, Chiang, Fang, Han, Wang, Hora, He, Lučić, Changpinyo, Tripathi, Youssef, Kwak, Schlattner,
  Graves, Leblond, Zeng, Andreassen, Rasskin, Song, Cao, Oh, Hoffman, Skut, Zhang, Stritar, Cai, Khanna, Wang, Sharma, Reisswig, Jun, Prasad, Sholokhova, Singh, Rosenthal, Ruoss, Beaufays, Kirmani, Chen, Schalkwyk, Herzig, Kim, Jacob, Vincent, Reyes, Balazevic, Hussenot, Schneider, Barnes, Castro, Babbula, Green, Cabi, Duduta, Driess, Galt, Velan, Wang, Jiao, Mauger, Phan, Patel, Galić, Chang, Marcus, Harvey, Salazar, Dabir, Sheth, Mandhane, Sedghi, Willcock, Zandieh, Prabhakara, Amini, Miech, Stone, Nicosia, Niemczyk, Xiao, Kim, Kwasiborski, Verma, Oflazer, Hirnschall, Sung, Liu, Everett, Bakker, Ágoston Weisz, Wang, Sampathkumar, Shaham, Xu, Altun, Wang, Saeki, Chen, Taropa, Vasanth, Austin, Huang, Petrovic, Dou, Golovin, Rozhdestvenskiy, Culp, Wu, Sano, Jain, Proskurnia, Cevey, Ruiz, Patil, Mirzazadeh, Ni, Snaider, Fan, Fréchette, Pierigiovanni, Iqbal, Lee, Fantacci, Xing, Wang, Irpan, Raposo, Luan, Chen, Ganapathy, Hui, Nie, Guyon, Ge, Vij, Zheng, Lee, Castaño, Baatarsukh, Ibagon, Chronopoulou,
  FitzGerald, Viswanadha, Huda, Moroshko, Stoyanov, Kolhar, Vaucher, Watts, Kuncoro, Michalewski, Kambala, Batsaikhan, Andreev, Jurenka, Le, Chen, Jishi, Chakera, Chen, Kini, Yadav, Siddhant, Labzovsky, Lakshminarayanan, Bostock, Botadra, Anand, Bishop, Conway-Rahman, Agarwal, Donchev, Singhal, de~Chaumont~Quitry, Ponomareva, Agrawal, Ni, Krishna, Samsikova, Karro, Du, von Glehn, Lu, Choquette-Choo, Qin, Zhang, Li, Tyam, Mishra, Lowe, Ji, Wang, Faruqui, Slone, Dalibard, Narayanaswamy, Lambert, Manzagol, Karliner, Bolt, Lobov, Kusupati, Ye, Yang, Zen, George, Bhutani, Lacombe, Riachi, Bansal, Soh, Gao, Yu, Yu, Nottage, Rojas-Esponda, Noraky, Gupta, Kotikalapudi, Chang, Deur, Graur, Mossin, Farnese, Figueira, Moufarek, Huang, Zochbauer, Ingram, Chen, Wu, Puigdomènech, Rechis, Yu, Padmanabhan, Zhu, ling Ko, Banino, Daruki, Selvan, Bhaswar, Diaz, Su, Scellato, Brennan, Han, Chung, Agrawal, Khandelwal, Sim, Lustman, Ritter, Guu, Xia, Jain, Wang, Hill, Rossini, Kostelac, Misiunas, Sabne, Kim, Iscen, Wang, Leal,
  Sreevatsa, Evci, Warmuth, Joshi, Suo, Lottes, Honke, Jou, Karp, Hu, Sahni, Taïga, Kong, Ghosh, Wang, Pavagadhi, Axelsson, Grigorev, Siegler, Lin, Wang, Parisotto, Maddineni, Subudhi, Ben-David, Pochernina, Keller, Avrahami, Yuan, Mehta, Liu, Yang, Kan, Lee, Funkhouser, Cheng, Shi, Sharma, Kelley, Eyal, Malkov, Tallec, Bahat, Yan, Xintian, Wu, Lindner, Wu, Caciularu, Luo, Jenatton, Zaman, Bi, Kornakov, Mallya, Ikeda, Karo, Singh, Evans, Netrapalli, Nallatamby, Tian, Assael, Raunak, Carbune, Bica, Madmoni, Cattle, Grover, Somandepalli, Lall, Vázquez-Reina, Patana, Mu, Talluri, Tran, Aggarwal, Skerry-Ryan, Xu, Burrows, Pan, Yvinec, Lu, Zhang, Nguyen, Mu, Barcik, Ran, Beltrone, Choromanski, Kharrat, Albanie, Purser-haskell, Bieber, Zhang, Wang, Hudson, Zhang, Fu, Mauerer, Bateni, Maschinot, Wang, Zhu, Pillai, Weyand, Liu, Akerlund, Bertsch, Premachandran, Jin, Roulet, de~Boursac, Mittal, Ndebele, Karadzhov, Ghalebikesabi, Liang, Wu, Cong, Ghelani, Singh, Fatemi, Warren, Chen, Kwong, Kolganov, Li, Song, Kuang,
  Miryoosefi, Webster, Wendt, Socala, Su, Mendonça, Gupta, Li, Tsai, Qiong, Hu, Kang, Chen, Girgin, Xian, Lee, Ramsden, Baker, Elish, Krayvanova, Joshi, Simsa, Yang, Ambroszczyk, Ghosh, Kar, Shangguan, Yamamori, Akulov, Brock, Tang, Vashishtha, Munoz, Steiner, Andra, Eppens, Feng, Kobayashi, Goldshtein, Mahdy, Wang, Jilei, Wang, Killam, Kwiatkowski, Kopparapu, Zhan, Jia, Bendebury, Luo, Recasens, Knight, Chen, Patel, Li, Withbroe, Weesner, Bhatia, Ren, Eisenbud, Songhori, Sun, Choma, Kementsietsidis, Manning, Roark, Farhan, Feng, Tatineni, Cobon-Kerr, Li, Hendricks, Noble, Breaux, Kushman, Peng, Xue, Tobin, Rogers, Lipschultz, Alberti, Vlaskin, Dehghani, Sharma, Warkentin, Lee, Uria, Juan, Chandorkar, Sheftel, Liu, Davoodi, Pigem, Dhamdhere, Ross, Hoech, Mahdieh, Liu, Li, McCafferty, Liu, Mircea, Song, Savant, Saade, Cherry, Hellendoorn, Goyal, Pucciarelli, Torres, Yahav, Lee, Sjoesund, Kirov, Chang, Ghoshal, Li, Baechler, Pereira, Sainath, Boral, Grewe, Halumi, Phu, Shen, Ribeiro, Varma, Kaskasoli,
  Feinberg, Potti, Kahn, Wisniewski, Mohamed, Hrafnkelsson, Shahriari, Lespiau, Patel, Yeung, Paine, Mei, Ramirez, Shivanna, Zhong, Woodward, Tubone, Khan, Chen, Nielsen, Ionescu, Prabhu, Gao, Wang, Augenstein, Subramaniam, Chang, Iliopoulos, Luo, Khan, Kuo, Teplyashin, Perot, Kilpatrick, Globerson, Yu, Siddiqui, Sukhanov, Kandoor, Gupta, Andreetto, Ambar, Kim, Wesołowski, Perrin, Limonchik, Fan, Stephan, Stewart-Binks, Kappedal, He, Cogan, Datta, Zhou, Ye, Kieliger, Ramalho, Kastner, Mentzer, Ko, Suggala, Zhou, Butt, Strejček, Belenki, Venugopalan, Ling, Eltyshev, Deng, Kovacs, Raghavachari, Dai, Schuster, Schwarcz, Nguyen, Nguyen, Buttimore, Mallick, Gandhe, Benjamin, Jastrzebski, Yan, Basu, Apps, Edkins, Allingham, Odisho, Kocisky, Zhao, Xue, Reddy, Anastasiou, Atias, Redmond, Milan, Heess, Schmit, Dafoe, Andor, Gangwani, Dragan, Zhang, Kachra, Wu, Xue, Aydin, Liu, Zhou, Malihi, Wu, Gopal, Schumann, Stys, Wang, Olšák, Liu, Schallhart, Mao, Brady, Xu, Mery, Sitawarin, Velusamy, Cobley, Zhai, Walder,
  Katz, Jawahar, Kulkarni, Yang, Paszke, Wang, Damoc, Borsos, Smith, Li, Gupta, Kapishnikov, Prakash, Luisier, Agarwal, Grathwohl, Chen, Han, Mehta, Over, Azizi, Meng, Santo, Zheng, Shapiro, Petrovski, Hui, Ghafouri, Snoek, Qin, Jordan, Sikora, Malmaud, Kuang, Świetlik, Sang, Shi, Li, Rosenberg, Zhao, Crawford, Peter, Lei, Garcia, Le, Wang, Amelot, Orr, Kacham, Alon, Tyen, Arora, Lyon, Kurakin, Ly, Guidroz, Yan, Panigrahy, Xu, Kagohara, Cheng, Noland, Lee, Lee, Yip, Wang, Nehoran, Bykovsky, Shan, Bhagatwala, Yan, Tan, Garrido, Ethier, Hurley, Vesom, Chen, Qiao, Nayyar, Walker, Sandhu, Rosca, Swisher, Dektiarev, Dillon, Muraru, Tragut, Myaskovsky, Reid, Velic, Xiao, George, Brand, Li, Yu, Gu, Deng, Aubet, Yeganeh, Alcober, Smith, Cohn, McKinney, Tschannen, Sampath, Cheon, Luo, Liu, Orbay, Peng, Botea, Zhang, Yoon, Magalhaes, Stradomski, Mackinnon, Hemingray, Venkatesan, May, Kim, Druinsky, Ye, Xu, Huang, Abdallah, Dostmohamed, Fellinger, Munkhdalai, Maurya, Garst, Zhang, Krikun, Bucher, Veerubhotla, Liu, Li,
  Gupta, Adamek, Chen, Orlando, Zaks, van Amersfoort, Camp, Wan, Choe, Wu, Olszewska, Yu, Vadali, Scholz, Freitas, Lin, Hua, Liu, Ding, Zhou, Severson, Tsihlas, Yang, Spalink, Yerram, Pankov, Blevins, Vargas, Jauhari, Miecnikowski, Zhang, Kumar, Farabet, Lan, Flennerhag, Bitton, Ma, Bražinskas, Collins, Ahuja, Kudugunta, Bortsova, Giang, Zhu, Chi, Lundberg, Stern, Puttagunta, Xiong, Wu, Pande, Jhindal, Murphy, Clark, Brockschmidt, Deines, McKee, Bahir, Shen, Truong, McDuff, Gesmundo, Rosseel, Liang, Caluwaerts, Hamrick, Kready, Cassin, Ingale, Lao, Pollom, Ding, He, Bellot, Iljazi, Boppana, Han, Thompson, Khalifa, Bulanova, Mitrevski, Pang, Cooney, Shi, Coaguila, Yakar, Ranzato, Momchev, Rawles, Charles, Maeng, Zhang, Bansal, Zhao, Albert, Yuan, Vijayanarasimhan, Hirsch, Ramasesh, Vodrahalli, Wang, Gupta, Strouse, Ni, Patel, Taubman, Huo, Gharibian, Monteiro, Lam, Vasudevan, Chaudhary, Albuquerque, Gupta, Riedel, Hegde, Ruderman, György, Wainwright, Chaugule, Ayan, Levinboim, Shleifer, Kalley, Mirrokni,
  Rao, Radhakrishnan, Hartford, Wu, Zhu, Bertolini, Xiong, Serrano, Tomlinson, Ott, Chang, Graham, Li, Liang, Long, Borgeaud, Ahmad, Grills, Mincu, Izzard, Liu, Xie, O'Bryan, Ponda, Tong, Liu, Malkin, Salama, Chen, Anil, Rao, Swavely, Bilenko, Anderson, Tan, Xie, Wu, Yu, Vinyals, Ryabtsev, Dangovski, Baumli, Keysers, Wright, Ashwood, Chan, Shtefan, Guo, Bapna, Soricut, Pecht, Ramos, Wang, Cai, Trinh, Barham, Friso, Stickgold, Ding, Shakeri, Ardila, Briakou, Culliton, Raveret, Cui, Saxton, Roy, Azizi, Yin, Loher, Bunner, Choi, Ahmed, Li, Li, Dai, Elabd, Ganapathy, Agrawal, Hua, Kunkle, Rajayogam, Ahuja, Conmy, Vasiloff, Beak, Yew, Mudigonda, Wydrowski, Blanton, Wang, Dauphin, Xu, Polacek, Chen, Hu, Sho, Kunesch, Manshadi, Rutherford, Li, Hsiao, Barr, Tudor, Kecman, Nagrani, Pchelin, Sundermeyer, S, Karmarkar, Gao, Chole, Bachem, Gao, BC, Dibb, Verzetti, Hernandez-Campos, Lunts, Johnson, Trapani, Koster, Brusilovsky, Xiong, Mohabey, Ke, Zou, Sabolić, Campos, Palowitch, Morris, Qiu, Ponnuramu, Li, Sharma,
  Sodhia, Tekelioglu, Chuklin, Yenugula, Gemzer, Strinopoulos, El-Husseini, Wang, Zhong, Leurent, Natsev, Wang, Mahaarachchi, Zhu, Peng, Alabed, Lee, Brohan, Szlam, Oh, Kovsharov, Lee, Wong, Barnes, Thornton, Gimeno, Levy, Sevenich, Johnson, Mallinson, Dadashi, Wang, Ren, Lahoti, Dhar, Feldman, Zheng, Ulrich, Panait, Blokzijl, Baetu, Matak, Harlalka, Shah, Marian, von Dincklage, Du, Ley-Wild, Brownfield, Schumacher, Stuken, Noghabi, Gupta, Ren, Malmi, Weissenberger, Huergo, Bauza, Lampe, Douillard, Seyedhosseini, Frostig, Ghahramani, Nguyen, Krishnakumar, Ye, Gupta, Nazari, Geirhos, Shaw, Eleryan, Damen, Palomaki, Xiao, Wu, Yuan, Meadowlark, Bilotti, Lin, Sridhar, Schroecker, Chung, Luo, Strohman, Liu, Zheng, Emond, Wang, Lampinen, Fukuzawa, Campbell-Ajala, Roy, Lee-Thorp, Wang, Naim, Tony, ên, Bensky, Gupta, Rogozińska, Fu, Pillai, Veličković, Drath, Neubeck, Tulsyan, Klimovskiy, Metzler, Stevens, Yeh, Yuan, Yu, Zhang, Go, Tsang, Xu, Wan, Galatzer-Levy, Sobell, Toki, Salesky, Zhou, Antognini, Douglas,
  Wu, Lelkes, Kim, Cavallaro, Salazar, Liu, Besley, Refice, Jia, Li, Sokolik, Kannan, Simon, Chick, Aharon, Gandhi, Daswani, Amiri, Birodkar, Ittycheriah, Grabowski, Chang, Sutton, Zhixin, Lai, Telang, Sargsyan, Jiang, Hoffmann, Brichtova, Hessel, Halcrow, Jerome, Brown, Tomala, Buchatskaya, Yu, Menon, Moreno, Liao, Zayats, Tang, Mah, Shenoy, Siegman, Hadian, Kwon, Tu, Khajehnouri, Foley, Haghani, Wu, Keshava, Gupta, Bruguier, Yao, Karmon, Zintgraf, Wang, Piqueras, Jung, Brennan, Machado, Giustina, Tessler, Lee, Zhang, Moore, Daugaard, Frömmgen, Beattie, Zhang, Kasenberg, Geri, Qin, Tomar, Ouyang, Yu, Zhou, Mathews, Davis, Li, Gupta, Yates, Deng, Kemp, Joung, Vassilvitskii, Guo, LV, Dopson, Lachgar, McConnaughey, Choudhury, Dena, Cohen, Ainslie, Levi, Gopavarapu, Zablotskaia, Vallet, Bahargam, Tang, Tomasev, Dyer, Balle, Lee, Bono, Mendez, Zubov, Yang, Rendulic, Zheng, Hogue, Pundak, Leith, Bhoopchand, Han, Žanić, Schaul, Delakis, Iyer, Wang, Singh, Abdelhamed, Thomas, Brahma, Dib, Kumar, Zhou, Bai,
  Mishra, Sun, Anklin, Sukkerd, Agubuzu, Briukhov, Gulati, Sieb, Pardo, Nasso, Chen, Zhu, Sosea, Goldin, Rush, Hombaiah, Noever, Zhou, Haves, Phuong, Ades, ting Chen, Yang, Pagadora, Bileschi, Cotruta, Saputro, Pramanik, Ammirati, Garrette, Villela, Blyth, Akbulut, Jha, Rrustemi, Wongpanich, Nagpal, Wu, Rivière, Kishchenko, Srinivasan, Chen, Sinha, Pham, Jia, Hennigan, Bakalov, Attaluri, Garmon, Rodriguez, Wegner, Jia, Senter, Fiedel, Petek, Liu, Hardin, Lehri, Carreira, Smoot, Prasetya, Akazawa, Stefanoiu, Ho, Angelova, Lin, Kim, Chen, Sieniek, Li, Guo, Baltateanu, Tafti, Wunder, Olmert, Shukla, Shen, Kovelamudi, Venkatraman, Neel, Thoppilan, Connor, Benzing, Stjerngren, Ghiasi, Polozov, Howland, Weber, Chiu, Girirajan, Terzis, Wang, Li, Shalom, Tewari, Denton, Aharoni, Kalb, Zhao, Zhang, Filos, Rahtz, Jain, Fan, Rodrigues, Wang, Shin, Austin, Ring, Sanchez-Vargas, Hassen, Kessler, Alon, Zhang, Chen, Ma, Si, Hou, Mirhoseini, Wilson, Bacon, Roelofs, Shu, Vasudevan, Adler, Dwornik, Terzi, Lawlor, Askham,
  Bernico, Dong, Hidey, Kilgour, Liu, Bhupatiraju, Leonhard, Zuo, Talukdar, Wei, Severyn, Listík, Lee, Tripathi, Park, Matias, Liu, Ruiz, Jayaram, Tolins, Marcenac, Wang, Seybold, Prior, Sharma, Weber, Sirotenko, Sung, Du, Pavlick, Zinke, Freitag, Dylla, Arenas, Potikha, Goldman, Tao, Chhaparia, Voitovich, Dogra, Ražnatović, Tsai, You, Johnson, Tucker, Gu, Yoo, Majzoubi, Gabeur, Raad, Rhodes, Kolipaka, Howard, Sampemane, Li, Asawaroengchai, Nguyen, Zhang, Cour, Yu, Fu, Jiang, Huang, Surita, Iturrate, Karov, Collins, Baeuml, Fuchs, Shetty, Ramaswamy, Ebrahimi, Guo, Shar, Barth-Maron, Addepalli, Richter, Cheng, Rives, Zheng, Griesser, Dikkala, Zeldes, Safarli, Das, Srivastava, Khan, Li, Pandey, Markeeva, Belov, Yan, Rybiński, Chen, Nawhal, Quinn, Govindaraj, York, Roberts, Garg, Godbole, Abernethy, Das, Thiet, Tompson, Nham, Vats, Caine, Helmholz, Pongetti, Ko, An, Hu, Ling, Pawar, Leland, Kinoshita, Khawaja, Selvi, Ie, Sinopalnikov, Proleev, Tripuraneni, Bevilacqua, Lee, Sanford, Suh, Tran, Dean,
  Baumgartner, Heitkaemper, Gubbi, Toutanova, Xu, Thekkath, Rong, Jain, Xie, Virin, Li, Litchev, Powell, Bharti, Kraft, Hua, Ikonomidis, Hitron, Kumar, Matthey, Bridgers, Lax, Malhi, Skopek, Gupta, Cao, Rasquinha, Põder, Stokowiec, Roth, Li, Sander, Kessinger, Jain, Loper, Park, Yarom, Cheng, Guruganesh, Rao, Li, Barros, Sushkov, Ferng, Shah, Aharoni, Kumar, McConnell, Li, Wang, Pereira, Swanson, Jamil, Xiong, Vijayakumar, Shroff, Soparkar, Gu, Soares, Wang, Majmundar, Wei, Bailey, Kassner, Kawamoto, Žužić, Gomes, Gupta, Guzman, Dasgupta, Bai, Pan, Piccinno, Vogel, Ponce, Hutter, Chang, Jiang, Gog, Ionescu, Manyika, Pedregosa, Ragan, Behrman, Mullins, Devin, Pyne, Gawde, Chadwick, Gu, Tavakkol, Twigg, Goyal, Elue, Goldie, Venkatachary, Fei, Feng, Ritter, Leal, Dasari, Sun, Rochman, O'Donoghue, Liu, Sproch, Chen, Clay, Petrov, Sidhwani, Mihailescu, Panagopoulos, Piergiovanni, Bai, Powell, Karkhanis, Yacovone, Mitrichev, Kovac, Uthus, Yazdanbakhsh, Amos, Zheng, Zhang, Miao, Ramabhadran, Radpour, Thakoor,
  Newlan, Lang, Jankowski, Bharadwaj, Sarr, Ashraf, Mondal, Yan, Rawat, Velury, Kochanski, Eccles, Och, Sharma, Mahintorabi, Gurney, Muir, Cohen, Thakur, Bloniarz, Mujika, Pritzel, Caron, Rahman, Lang, Onoe, Sirkovic, Hoover, Jian, Duque, Narayanan, Soergel, Haig, Maggiore, Buch, Dean, Figotin, Karpov, Gupta, Zhou, Huang, Vaswani, Semturs, Shivakumar, Watanabe, Rajendran, Lu, Hou, Ye, Vashishth, Nti, Sakenas, Ni, DeCarlo, Bendersky, Bagri, Cano, Peake, Tokumine, Godbole, Guía, Lando, Selo, Ellis, Tarlow, Gillick, Epasto, Jonnalagadda, Wei, Xie, Taly, Paganini, Sundararajan, Toyama, Yu, Petrova, Pappu, Agrawal, Buthpitiya, Frye, Buschmann, Crocker, Tagliasacchi, Wang, Huang, Perel, Wieder, Kazawa, Wang, Cole, Gupta, Golan, Bang, Kulkarni, Franko, Liu, Reid, Dalmia, Whang, Cen, Sundaram, Ferret, Isik, Ionita, Sun, Shekhawat, Mohammad, Pham, Huang, Raman, Zhou, Mcilroy, Myers, Peng, Scott, Covington, Erell, Joshi, Oliveira, Noy, Nasir, Walker, Axelrod, Dozat, Han, Chu, Weinstein, Shukla, Chandrakaladharan,
  Poklukar, Li, Jin, Eruvbetine, Hansen, Dabush, Jacovi, Phatale, Zhu, Baker, Shomrat, Xiao, Pouget-Abadie, Zhang, Wei, Song, King, Huang, Zhu, Sun, Franco, Lin, Arora, Hui, Li, Xia, Vilnis, Schain, Alarakyia, Prince, Phillips, Habtegebriel, Xu, Gui, Ontanon, Aroyo, Gill, Lu, Katariya, Madeka, Krishnan, Raghvendra, Freedman, Tay, Menghani, Choy, Shetty, Abolafia, Kukliansky, Chou, Lichtarge, Burke, Coleman, Guo, Jin, Bhattacharya, Langston, Li, Kotecha, Yakubovich, Chen, Petrov, Powell, He, Quick, Garg, Hwang, Lu, Bhojanapalli, Kjems, Mehran, Archer, van Hasselt, Balakrishna, Kearns, Guo, Riesa, Sazanovich, Gao, Sauer, Yang, Sheng, Jimma, Gansbeke, Nikolaev, Wei, Millican, Zhao, Snyder, Bolelli, O'Brien, Xu, Xia, Yuan, Neelakantan, Barker, Yadav, Kirkwood, Ahmad, Wee, Grimstad, Wang, Wiethoff, Settle, Wang, Blundell, Chen, Duvarney, Hu, Ronneberger, Lee, Li, Chakladar, Butryna, Evangelopoulos, Desjardins, Kanerva, Wang, Nowak, Li, Loo, Khurshudov, Shafey, Baddi, Lenc, Razeghi, Lieber, Sinha, Ma, Su, Huang,
  Ushio, Klimczak-Plucińska, Mohamed, Chen, Osindero, Ginzburg, Lamprou, Bashlovkina, Tran, Khodaei, Anand, Di, Eskander, Vuyyuru, Liu, Kamath, Goldenberg, Bellaiche, Pluto, Rosgen, Mansoor, Wong, Ganesh, Bailey, Baird, Deutsch, Baek, Jia, Lee, Friesen, Braun, Lee, Panda, Hernandez, Williams, Liu, Liang, Autef, Pitler, Jain, Kirk, Bunyan, Elias, Yin, Reid, Pope, Putikhin, Samanta, Guadarrama, Kim, Rowe, Valentine, Yan, Salcianu, Silver, Song, Singh, Ye, DeBalsi, Merey, Ofek, Webson, Mourad, Kakarla, Lattanzi, Roy, Sluzhaev, Butterfield, Tonioni, Waters, Kopalle, Chase, Cohan, Rao, Berry, Voznesensky, Hu, Chiafullo, Chikkerur, Scrivener, Zheng, Wiesner, Macherey, Lillicrap, Liu, Walker, Welling, Davies, Huang, Ren, Shabat, Agostini, Iinuma, Zelle, Sathyanarayana, D'olimpio, Redshaw, Ginsberg, Murthy, Geller, Matejovicova, Chakrabarti, Julian, Chan, Hu, Jarrett, Agarwal, Challagundla, Li, Tata, Ding, Meng, Dai, Vezzani, Garg, Bulian, Jasarevic, Cai, Rajamani, Santoro, Hartmann, Liang, Perz, Jindal, Bu, Seo,
  Poplin, Goedeckemeyer, Ghazi, Khadke, Liu, Mather, Zhang, Shah, Chen, Wei, Shivam, Cao, Cho, Scarpati, Moffitt, Barbu, Jurin, Chang, Liu, Zheng, Dave, Kaeser-Chen, Yu, Abdagic, Gonzalez, Huang, Zhong, Schmid, Petrini, Wertheim, Zhu, Nguyen, Ji, Zhou, Zhou, Feng, Cohen, Rim, Phal, Georgiev, Brand, Ma, Li, Gupta, Wang, Dubov, Tarbouriech, Majumder, Li, Rink, Suman, Guo, Sun, Nair, Xu, Elhawaty, Cabrera, Han, Eisenschlos, Bai, Li, Bansal, Sellam, Khan, Nguyen, Mao-Jones, Parotsidis, Marcus, Fan, Zimmermann, Kochinski, Graesser, Behbahani, Caceres, Riley, Kane, Lefdal, Willoughby, Vicol, Wang, Zhang, Gill, Liang, Prasad, Mariooryad, Kazemi, Wang, Muralidharan, Voigtlaender, Zhao, Zhou, D'Souza, Mavalankar, Arnold, Young, Sarvana, Lee, Nasr, Zou, Kim, Haas, Patel, Bulut, Parkinson, Biles, Kalashnikov, To, Kumar, Austin, Greve, Zhang, Goel, Li, Yaroshenko, Chang, Jindal, Clark, Taitelbaum, Johnson, Roval, Ko, Mohananey, Schuler, Dodhia, Li, Osawa, Cui, Xu, Shah, Huang, Gruzewska, Clement, Verma, Sercinoglu, Qian,
  Shah, Yamaguchi, Modi, Kosakai, Strohmann, Zeng, Gunel, Qian, Tarango, Jastrzębski, David, Shan, Schuh, Lad, Gierke, Madhavan, Chen, Kurzeja, Santamaria-Fernandez, Chen, Cordell, Chervonyi, Garcia, Kannen, Perot, Ding, Cohen-Ganor, Lavrenko, Wu, Evans, dos Santos, Sewak, Brown, Hard, Puigcerver, Zheng, Liang, Gladchenko, Ingle, First, Sermanet, Magister, Velimirović, Reddi, Ricco, Agustsson, Adam, Levine, Gaddy, Holtmann-Rice, Wang, Sathe, Roy, Bratanič, Carin, Mehta, Bonacina, Cao, Finkelstein, Rieser, Wu, Altché, Scandinaro, Li, Vieillard, Sethi, Tanzer, Xing, Wang, Bhatia, Citovsky, Anthony, Lin, Shi, Jakobovits, Gibson, Apte, Lee, Chen, Byravan, Maniatis, Webster, Dai, Chen, Pan, Fadeeva, Gleicher, Luong, and Bhumihar}]{comanici2025gemini25pushingfrontier}
Gheorghe Comanici, Eric Bieber, Mike Schaekermann, Ice Pasupat, Noveen Sachdeva, Inderjit Dhillon, Marcel Blistein, Ori Ram, Dan Zhang, Evan Rosen, Luke Marris, Sam Petulla, Colin Gaffney, Asaf Aharoni, Nathan Lintz, Tiago~Cardal Pais, Henrik Jacobsson, Idan Szpektor, Nan-Jiang Jiang, and 3290 others. 2025.
\newblock \href {https://arxiv.org/abs/2507.06261} {Gemini 2.5: Pushing the frontier with advanced reasoning, multimodality, long context, and next generation agentic capabilities}.
\newblock \emph{Preprint}, arXiv:2507.06261.

\bibitem[{Daniel~Han and team(2023)}]{unsloth}
Michael~Han Daniel~Han and Unsloth team. 2023.
\newblock \href {http://github.com/unslothai/unsloth} {Unsloth}.

\bibitem[{DeepSeek-AI et~al.(2025)DeepSeek-AI, Guo, Yang, Zhang, Song, Zhang, Xu, Zhu, Ma, Wang, Bi, Zhang, Yu, Wu, Wu, Gou, Shao, Li, Gao, Liu, Xue, Wang, Wu, Feng, Lu, Zhao, Deng, Zhang, Ruan, Dai, Chen, Ji, Li, Lin, Dai, Luo, Hao, Chen, Li, Zhang, Bao, Xu, Wang, Ding, Xin, Gao, Qu, Li, Guo, Li, Wang, Chen, Yuan, Qiu, Li, Cai, Ni, Liang, Chen, Dong, Hu, Gao, Guan, Huang, Yu, Wang, Zhang, Zhao, Wang, Zhang, Xu, Xia, Zhang, Zhang, Tang, Li, Wang, Li, Tian, Huang, Zhang, Wang, Chen, Du, Ge, Zhang, Pan, Wang, Chen, Jin, Chen, Lu, Zhou, Chen, Ye, Wang, Yu, Zhou, Pan, Li, Zhou, Wu, Ye, Yun, Pei, Sun, Wang, Zeng, Zhao, Liu, Liang, Gao, Yu, Zhang, Xiao, An, Liu, Wang, Chen, Nie, Cheng, Liu, Xie, Liu, Yang, Li, Su, Lin, Li, Jin, Shen, Chen, Sun, Wang, Song, Zhou, Wang, Shan, Li, Wang, Wei, Zhang, Xu, Li, Zhao, Sun, Wang, Yu, Zhang, Shi, Xiong, He, Piao, Wang, Tan, Ma, Liu, Guo, Ou, Wang, Gong, Zou, He, Xiong, Luo, You, Liu, Zhou, Zhu, Xu, Huang, Li, Zheng, Zhu, Ma, Tang, Zha, Yan, Ren, Ren, Sha, Fu, Xu, Xie, Zhang,
  Hao, Ma, Yan, Wu, Gu, Zhu, Liu, Li, Xie, Song, Pan, Huang, Xu, Zhang, and Zhang}]{deepseekai2025deepseekr1incentivizingreasoningcapability}
DeepSeek-AI, Daya Guo, Dejian Yang, Haowei Zhang, Junxiao Song, Ruoyu Zhang, Runxin Xu, Qihao Zhu, Shirong Ma, Peiyi Wang, Xiao Bi, Xiaokang Zhang, Xingkai Yu, Yu~Wu, Z.~F. Wu, Zhibin Gou, Zhihong Shao, Zhuoshu Li, Ziyi Gao, and 181 others. 2025.
\newblock \href {https://arxiv.org/abs/2501.12948} {Deepseek-r1: Incentivizing reasoning capability in llms via reinforcement learning}.
\newblock \emph{Preprint}, arXiv:2501.12948.

\bibitem[{Goddard et~al.(2024)Goddard, Siriwardhana, Ehghaghi, Meyers, Karpukhin, Benedict, McQuade, and Solawetz}]{goddard-etal-2024-arcees}
Charles Goddard, Shamane Siriwardhana, Malikeh Ehghaghi, Luke Meyers, Vladimir Karpukhin, Brian Benedict, Mark McQuade, and Jacob Solawetz. 2024.
\newblock \href {https://doi.org/10.18653/v1/2024.emnlp-industry.36} {Arcee{'}s {M}erge{K}it: A toolkit for merging large language models}.
\newblock In \emph{Proceedings of the 2024 Conference on Empirical Methods in Natural Language Processing: Industry Track}, pages 477--485, Miami, Florida, US. Association for Computational Linguistics.

\bibitem[{Hendrycks et~al.(2021)Hendrycks, Burns, Kadavath, Arora, Basart, Tang, Song, and Steinhardt}]{hendrycks2021measuringmathematicalproblemsolving}
Dan Hendrycks, Collin Burns, Saurav Kadavath, Akul Arora, Steven Basart, Eric Tang, Dawn Song, and Jacob Steinhardt. 2021.
\newblock \href {https://arxiv.org/abs/2103.03874} {Measuring mathematical problem solving with the math dataset}.
\newblock \emph{Preprint}, arXiv:2103.03874.

\bibitem[{Hu et~al.(2021)Hu, Shen, Wallis, Allen-Zhu, Li, Wang, Wang, and Chen}]{hu2021loralowrankadaptationlarge}
Edward~J. Hu, Yelong Shen, Phillip Wallis, Zeyuan Allen-Zhu, Yuanzhi Li, Shean Wang, Lu~Wang, and Weizhu Chen. 2021.
\newblock \href {https://arxiv.org/abs/2106.09685} {Lora: Low-rank adaptation of large language models}.
\newblock \emph{Preprint}, arXiv:2106.09685.

\bibitem[{Kingma and Ba(2017)}]{kingma2017adammethodstochasticoptimization}
Diederik~P. Kingma and Jimmy Ba. 2017.
\newblock \href {https://arxiv.org/abs/1412.6980} {Adam: A method for stochastic optimization}.
\newblock \emph{Preprint}, arXiv:1412.6980.

\bibitem[{Lambert et~al.(2025)Lambert, Morrison, Pyatkin, Huang, Ivison, Brahman, Miranda, Liu, Dziri, Lyu, Gu, Malik, Graf, Hwang, Yang, Bras, Tafjord, Wilhelm, Soldaini, Smith, Wang, Dasigi, and Hajishirzi}]{lambert2025tulu3pushingfrontiers}
Nathan Lambert, Jacob Morrison, Valentina Pyatkin, Shengyi Huang, Hamish Ivison, Faeze Brahman, Lester James~V. Miranda, Alisa Liu, Nouha Dziri, Shane Lyu, Yuling Gu, Saumya Malik, Victoria Graf, Jena~D. Hwang, Jiangjiang Yang, Ronan~Le Bras, Oyvind Tafjord, Chris Wilhelm, Luca Soldaini, and 4 others. 2025.
\newblock \href {https://arxiv.org/abs/2411.15124} {Tulu 3: Pushing frontiers in open language model post-training}.
\newblock \emph{Preprint}, arXiv:2411.15124.

\bibitem[{Liu et~al.(2025{\natexlab{a}})Liu, Chen, Shoeybi, Catanzaro, and Ping}]{liu2025acemathadvancingfrontiermath}
Zihan Liu, Yang Chen, Mohammad Shoeybi, Bryan Catanzaro, and Wei Ping. 2025{\natexlab{a}}.
\newblock \href {https://arxiv.org/abs/2412.15084} {Acemath: Advancing frontier math reasoning with post-training and reward modeling}.
\newblock \emph{Preprint}, arXiv:2412.15084.

\bibitem[{Liu et~al.(2025{\natexlab{b}})Liu, Yang, Chen, Lee, Shoeybi, Catanzaro, and Ping}]{liu2025acereasonnemotron11advancingmath}
Zihan Liu, Zhuolin Yang, Yang Chen, Chankyu Lee, Mohammad Shoeybi, Bryan Catanzaro, and Wei Ping. 2025{\natexlab{b}}.
\newblock \href {https://arxiv.org/abs/2506.13284} {Acereason-nemotron 1.1: Advancing math and code reasoning through sft and rl synergy}.
\newblock \emph{Preprint}, arXiv:2506.13284.

\bibitem[{Mangrulkar et~al.(2022)Mangrulkar, Gugger, Debut, Belkada, Paul, and Bossan}]{peft}
Sourab Mangrulkar, Sylvain Gugger, Lysandre Debut, Younes Belkada, Sayak Paul, and Benjamin Bossan. 2022.
\newblock {PEFT}: State-of-the-art parameter-efficient fine-tuning methods.
\newblock \url{https://github.com/huggingface/peft}.

\bibitem[{Mnih et~al.(2016)Mnih, Badia, Mirza, Graves, Lillicrap, Harley, Silver, and Kavukcuoglu}]{mnih2016asynchronousmethodsdeepreinforcement}
Volodymyr Mnih, Adrià~Puigdomènech Badia, Mehdi Mirza, Alex Graves, Timothy~P. Lillicrap, Tim Harley, David Silver, and Koray Kavukcuoglu. 2016.
\newblock \href {https://arxiv.org/abs/1602.01783} {Asynchronous methods for deep reinforcement learning}.
\newblock \emph{Preprint}, arXiv:1602.01783.

\bibitem[{OpenAI(2024)}]{openai2024o1}
OpenAI. 2024.
\newblock \href {https://openai.com/o1} {Openai o1 model}.
\newblock Accessed: 2025-08-05.

\bibitem[{OpenAI(2025)}]{openai2025o3o4minirelease}
OpenAI. 2025.
\newblock \href {https://openai.com/index/introducing-o3-and-o4-mini/} {Introducing openai o3 and o4‑mini}.
\newblock Accessed: 2025‑08‑05.

\bibitem[{Shao et~al.(2024)Shao, Wang, Zhu, Xu, Song, Bi, Zhang, Zhang, Li, Wu, and Guo}]{shao2024deepseekmathpushinglimitsmathematical}
Zhihong Shao, Peiyi Wang, Qihao Zhu, Runxin Xu, Junxiao Song, Xiao Bi, Haowei Zhang, Mingchuan Zhang, Y.~K. Li, Y.~Wu, and Daya Guo. 2024.
\newblock \href {https://arxiv.org/abs/2402.03300} {Deepseekmath: Pushing the limits of mathematical reasoning in open language models}.
\newblock \emph{Preprint}, arXiv:2402.03300.

\bibitem[{Sheng et~al.(2024)Sheng, Zhang, Ye, Wu, Zhang, Zhang, Peng, Lin, and Wu}]{sheng2024hybridflow}
Guangming Sheng, Chi Zhang, Zilingfeng Ye, Xibin Wu, Wang Zhang, Ru~Zhang, Yanghua Peng, Haibin Lin, and Chuan Wu. 2024.
\newblock Hybridflow: A flexible and efficient rlhf framework.
\newblock \emph{arXiv preprint arXiv: 2409.19256}.

\bibitem[{Team et~al.(2025)Team, Du, Gao, Xing, Jiang, Chen, Li, Xiao, Du, Liao, Tang, Wang, Zhang, Yuan, Lu, Tang, Sung, Wei, Lai, Guo, Zhu, Ding, Hu, Yang, Zhang, Yao, Zhao, Lu, Li, Yu, Gao, Zheng, Yuan, Chen, Guo, Su, Wang, Zhao, Zhang, Liu, Yan, Wu, Shi, Ye, Yu, Dong, Zhang, Ma, Pan, Gong, Liu, Ma, Wei, Cao, Huang, Jiang, Gao, Xiong, He, Huang, Xu, Wu, He, Wei, Jia, Wu, Xu, Zu, Zhou, Pan, Charles, Li, Hu, Liu, Chen, Wang, Liu, Qin, Liu, Yang, Bao, Du, Wu, Wang, Zhou, Wang, Li, Zhu, Zhang, Wang, Yang, Huang, Huang, Xu, Yang, and Lin}]{kimiteam2025kimik15scalingreinforcement}
Kimi Team, Angang Du, Bofei Gao, Bowei Xing, Changjiu Jiang, Cheng Chen, Cheng Li, Chenjun Xiao, Chenzhuang Du, Chonghua Liao, Chuning Tang, Congcong Wang, Dehao Zhang, Enming Yuan, Enzhe Lu, Fengxiang Tang, Flood Sung, Guangda Wei, Guokun Lai, and 77 others. 2025.
\newblock \href {https://arxiv.org/abs/2501.12599} {Kimi k1.5: Scaling reinforcement learning with llms}.
\newblock \emph{Preprint}, arXiv:2501.12599.

\bibitem[{von Werra et~al.(2020)von Werra, Belkada, Tunstall, Beeching, Thrush, Lambert, Huang, Rasul, and Gallouédec}]{vonwerra2022trl}
Leandro von Werra, Younes Belkada, Lewis Tunstall, Edward Beeching, Tristan Thrush, Nathan Lambert, Shengyi Huang, Kashif Rasul, and Quentin Gallouédec. 2020.
\newblock Trl: Transformer reinforcement learning.
\newblock \url{https://github.com/huggingface/trl}.

\bibitem[{Wortsman et~al.(2022)Wortsman, Ilharco, Gadre, Roelofs, Gontijo-Lopes, Morcos, Namkoong, Farhadi, Carmon, Kornblith, and Schmidt}]{wortsman2022modelsoupsaveragingweights}
Mitchell Wortsman, Gabriel Ilharco, Samir~Yitzhak Gadre, Rebecca Roelofs, Raphael Gontijo-Lopes, Ari~S. Morcos, Hongseok Namkoong, Ali Farhadi, Yair Carmon, Simon Kornblith, and Ludwig Schmidt. 2022.
\newblock \href {https://arxiv.org/abs/2203.05482} {Model soups: averaging weights of multiple fine-tuned models improves accuracy without increasing inference time}.
\newblock \emph{Preprint}, arXiv:2203.05482.

\bibitem[{Wu et~al.(2025)Wu, Yao, Liu, Liu, Fu, Han, Li, Zhen, Zhong, and Yuan}]{wu2025unlockingefficientlongtoshortllm}
Han Wu, Yuxuan Yao, Shuqi Liu, Zehua Liu, Xiaojin Fu, Xiongwei Han, Xing Li, Hui-Ling Zhen, Tao Zhong, and Mingxuan Yuan. 2025.
\newblock \href {https://arxiv.org/abs/2503.20641} {Unlocking efficient long-to-short llm reasoning with model merging}.
\newblock \emph{Preprint}, arXiv:2503.20641.

\bibitem[{Yang et~al.(2024)Yang, Zhang, Hui, Gao, Yu, Li, Liu, Tu, Zhou, Lin, Lu, Xue, Lin, Liu, Ren, and Zhang}]{yang2024qwen25mathtechnicalreportmathematical}
An~Yang, Beichen Zhang, Binyuan Hui, Bofei Gao, Bowen Yu, Chengpeng Li, Dayiheng Liu, Jianhong Tu, Jingren Zhou, Junyang Lin, Keming Lu, Mingfeng Xue, Runji Lin, Tianyu Liu, Xingzhang Ren, and Zhenru Zhang. 2024.
\newblock \href {https://arxiv.org/abs/2409.12122} {Qwen2.5-math technical report: Toward mathematical expert model via self-improvement}.
\newblock \emph{Preprint}, arXiv:2409.12122.

\end{thebibliography}
\onecolumn
\appendix

\section{Prompt for Question Cleaning}
\label{app:filter}
The prompt for question cleaning is provided in Table \ref{tb: question_cleaning}

\section{Prompts for Answer Matching}
\label{app:eval}
The prompts for answer matching are provided in Table \ref{tb: answer_matching}

\section{Hyper-parameters for Supervised Fine Tuning}
\label{app:sft_hp}
The hyper-parameters for LoRA are provided in the Table \ref{tab:peft_config} and the hyper-parameters for SFT are provided in the Table \ref{tab:sft_config}.
\begin{table*}[h]
\centering
\begin{tabular}{ll}
\toprule
\textbf{Parameter} & \textbf{Value} \\
\midrule
Rank & 128 \\
LoRA Alpha & 128 \\
LoRA Dropout & 0.1 \\
Bias & none \\
Target Modules &
\texttt{\{q\_proj, k\_proj, v\_proj, o\_proj,} \\
& \texttt{gate\_proj, up\_proj, down\_proj, embeddings\}} \\
\bottomrule
\end{tabular}
\caption{PEFT configuration using LoRA.}
\label{tab:peft_config}
\vspace{1cm}
\begin{tabular}{ll}
\toprule
\textbf{Parameter} & \textbf{Value} \\
\midrule
Precision & bfloat16 \\
Max Sequence Length & 16,384 \\
Batch Size (per device) & 1 \\
Gradient Accumulation Steps & 16 \\
Effective Batch Size & 16 \\
Number of Epochs & 3 \\
Initial Learning Rate & $2 \times 10^{-5}$ \\
Final Learning Rate & $2 \times 10^{-7}$ \\
Learning Rate Scheduler & Linear \\
Optimizer & AdamW (8-bit) \\
Warmup Steps & 5 \\
Packing & False \\
Logging Steps & 1 \\
WandB Reporting & Enabled \\
\bottomrule
\end{tabular}
\caption{Training configuration used for supervised fine-tuning.}
\label{tab:sft_config}
\end{table*}

\section{Example Model Responses}
\label{app:examples}
\begin{figure}[h]
    \centering
    \includegraphics[width=\textwidth]{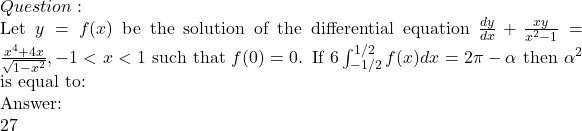}
    \caption{Sample question with the correct answer}
    \label{fg:ques}
\end{figure}

\begin{figure}[h]
    \centering
    \includegraphics[width=\textwidth]{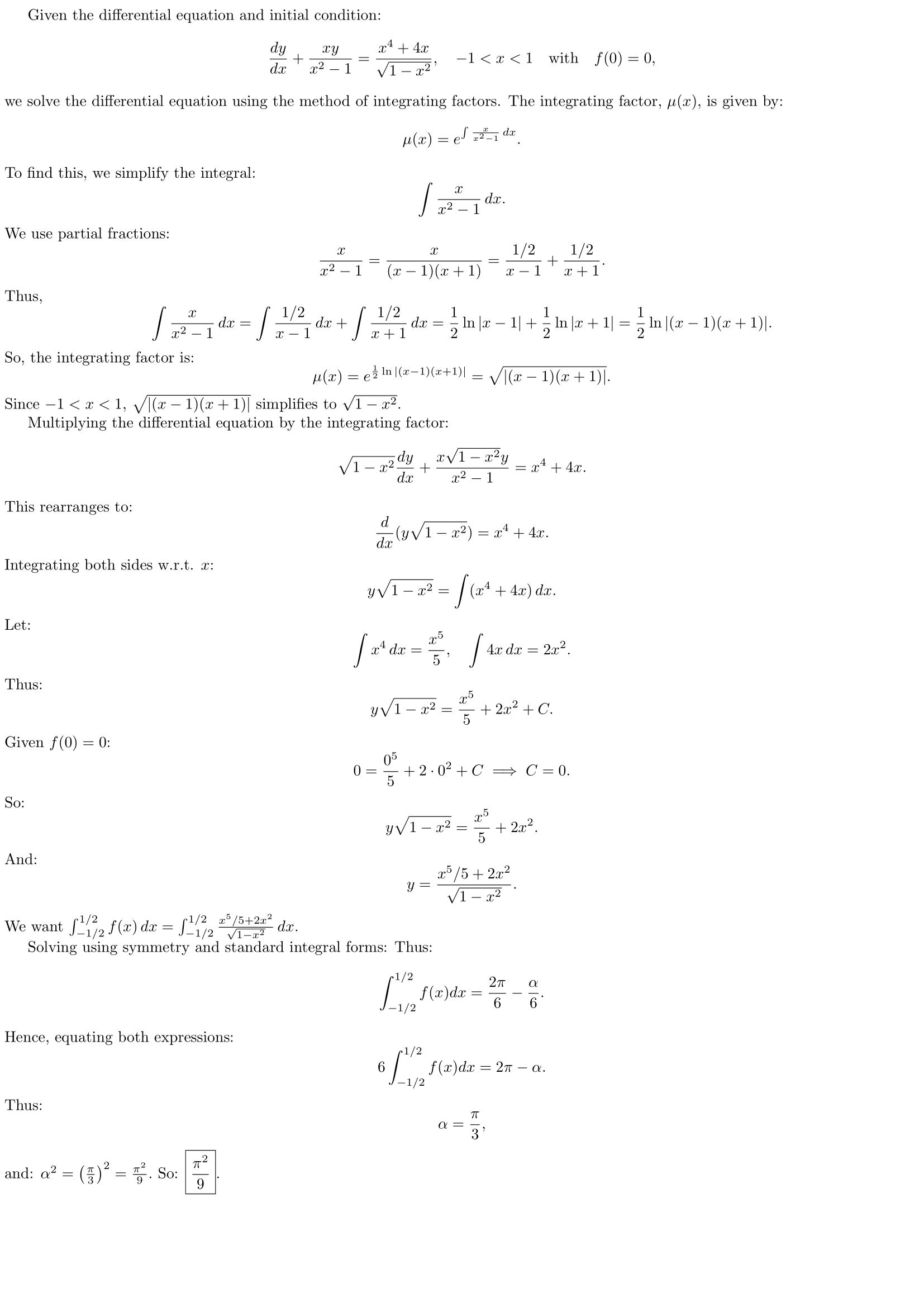}
    \caption{Response from GPT-4o (Part 1 of 1)}
    \label{fg:gpt1}
\end{figure}

\begin{figure}[h]
    \centering
    \includegraphics[width=\textwidth]{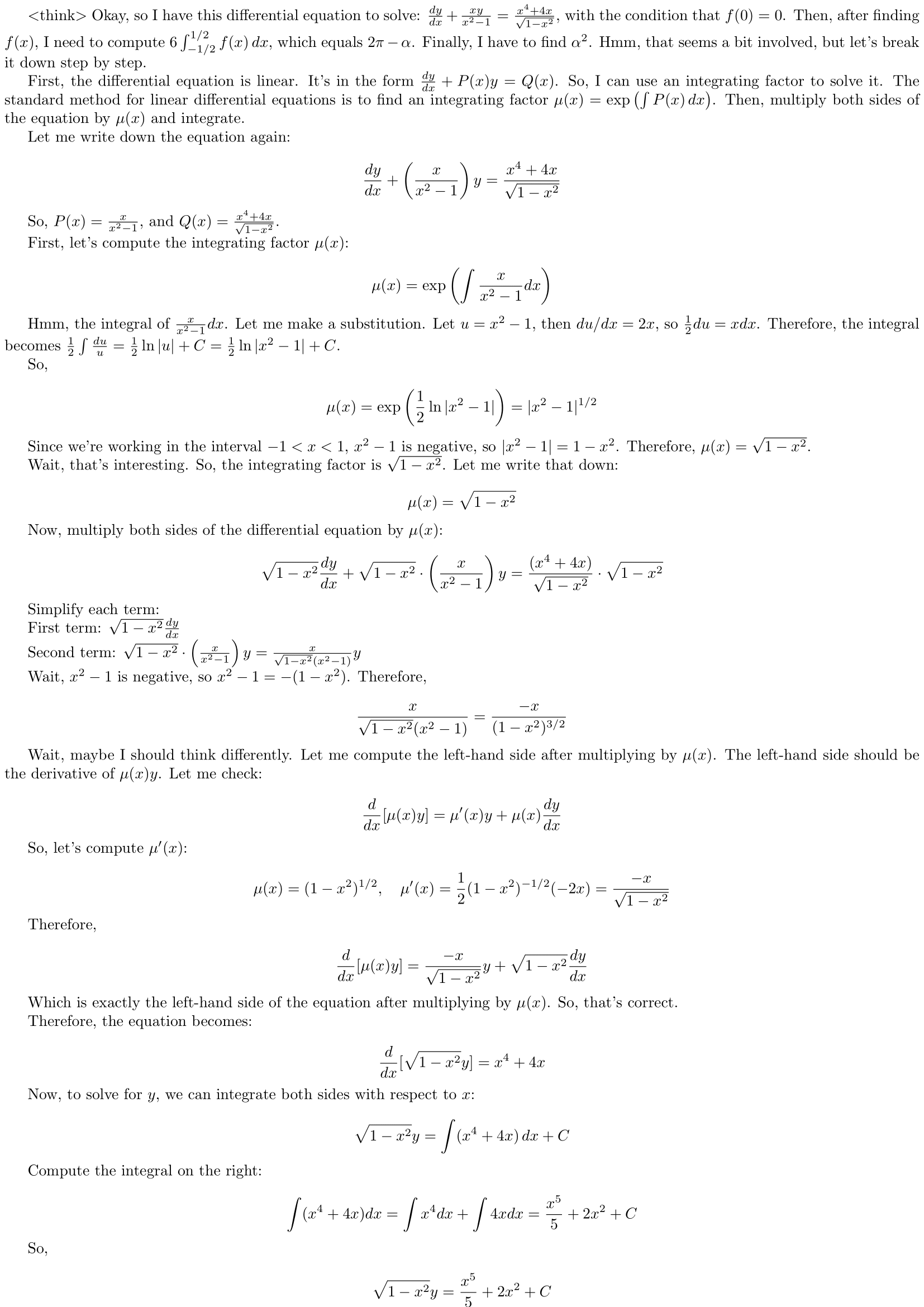}
    \caption{Response from DeepSeek R1 Distill Qwen 7B (Part 1 of 3)}
    \label{fg:r11}
\end{figure}

\begin{figure}[h]
    \centering
    \includegraphics[width=\textwidth]{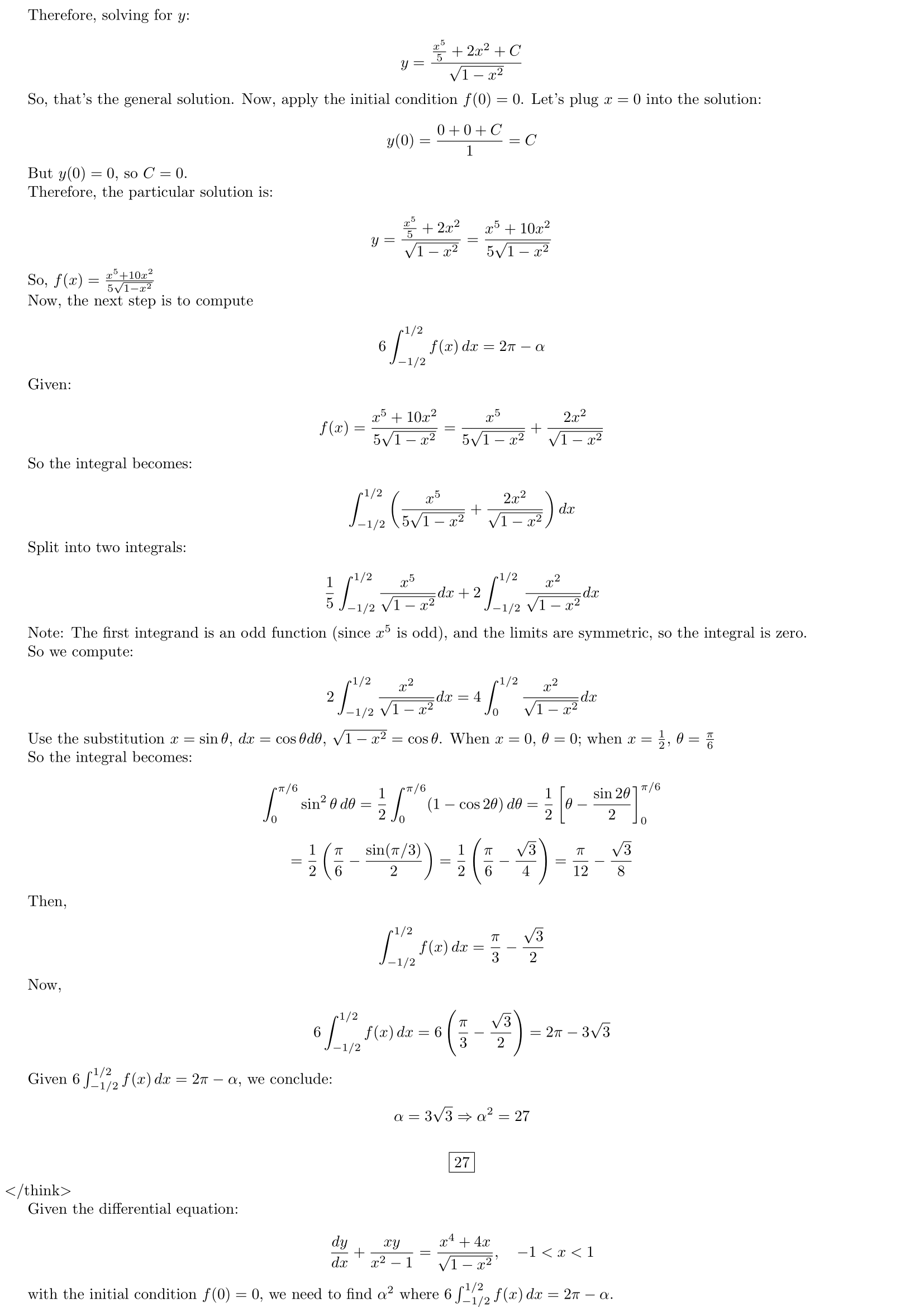}
    \caption{Response from DeepSeek R1 Distill Qwen 7B (Part 2 of 3)}
    \label{fg:r12}
\end{figure}

\begin{figure}[h]
    \centering
    \includegraphics[width=\textwidth]{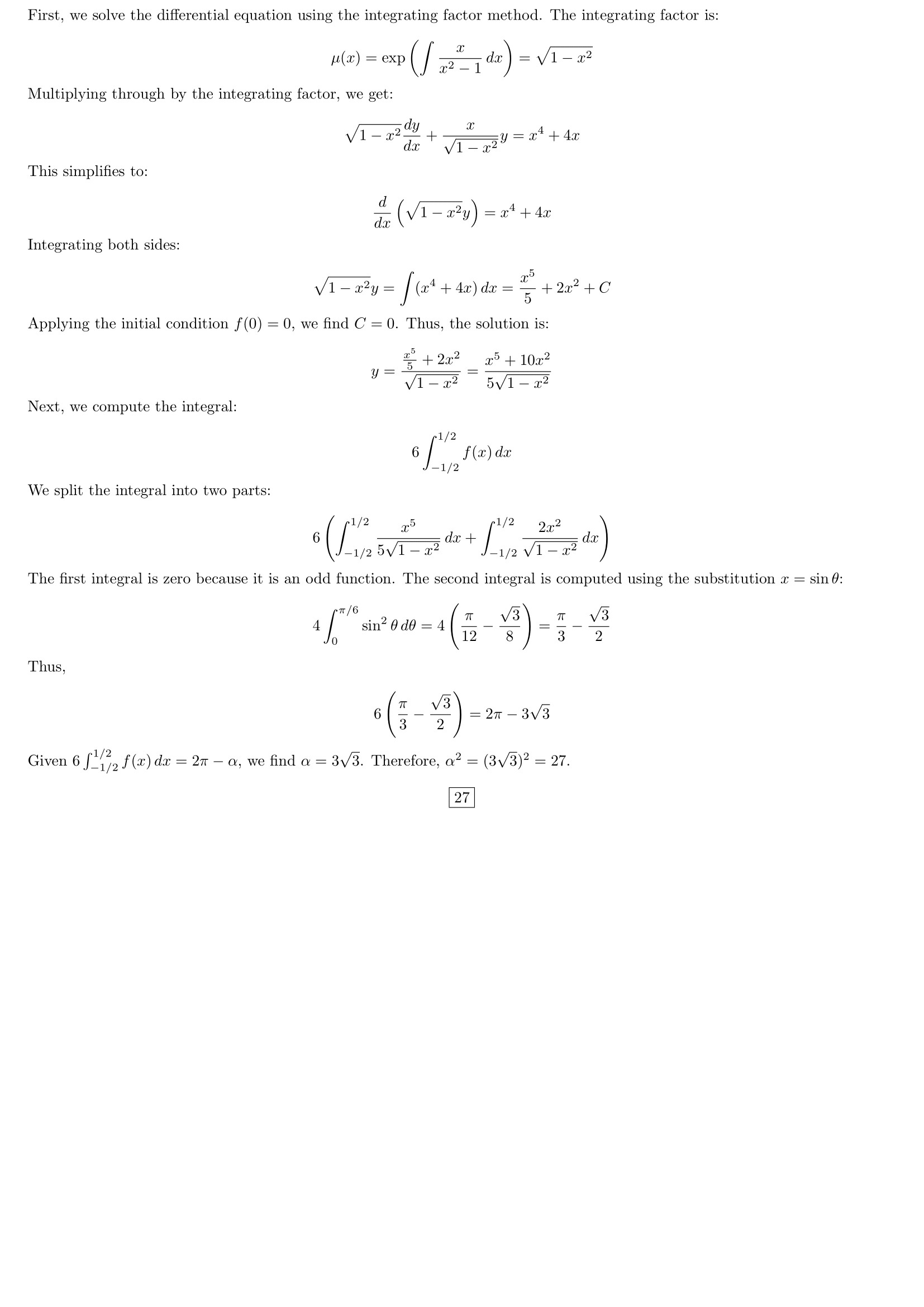}
    \caption{Response from DeepSeek R1 Distill Qwen 7B (Part 3 of 3)}
    \label{fg:r13}
\end{figure}

\begin{figure}[h]
    \centering
    \includegraphics[width=\textwidth]{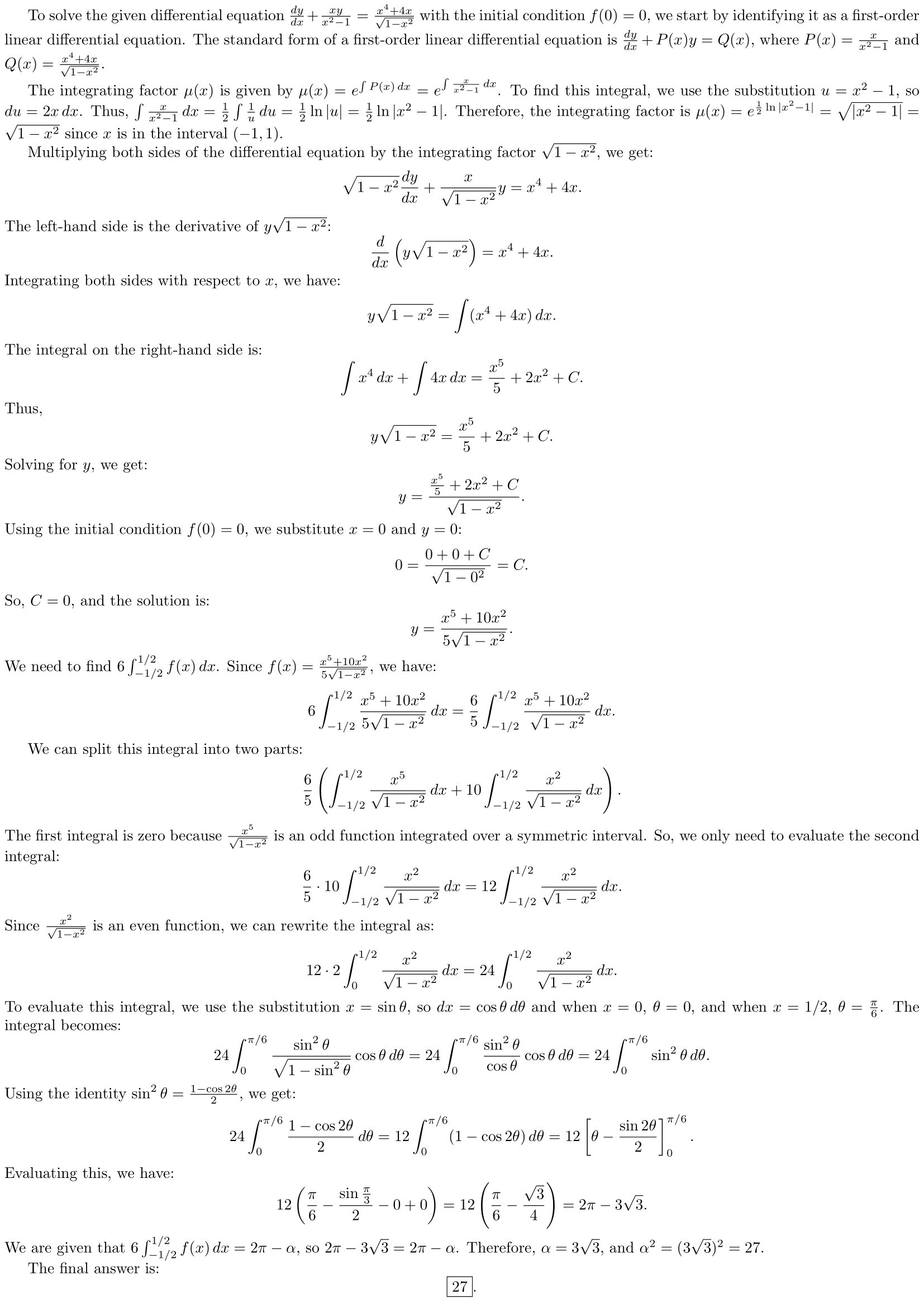}
    \caption{Response from Aryabhata 1.0 (Part 1 of 1)}
    \label{fg:arya1}
\end{figure}

The sample question along with its correct answer is presented in Figure~\ref{fg:ques}.  
\newline
The response generated by GPT-4o is shown in Figure~\ref{fg:gpt1}.  
\newline
The response produced by DeepSeek R1 Distill Qwen 7B is illustrated across Figures~\ref{fg:r11}, \ref{fg:r12}, and \ref{fg:r13}.  
\newline
The response from Aryabhata 1.0 is depicted in Figure~\ref{fg:arya1}.

\begin{table*}[t]
\begin{tabular}{|p{0.5\textwidth}|p{0.5\textwidth}|}
\hline
\textbf{MCQ} & \textbf{Numerical} \\
\hline
\textbf{System Prompt:} \newline
\begin{verbatim}
You are checking an MCQ. Given the list of 
options, determine if answer 1 and answer 2
are the same. Answer 1 is the same as answer
2 only if all the options match. Reason 
step-by-step and put the final answer YES 
or NO in \boxed{}. 
\end{verbatim}
&
\textbf{System Prompt:} \newline
\begin{verbatim}
You are checking an exam. For a given  
question, determine if answer 1 and answer 
2 are the same. Since the answers are for 
the same question, you can assume similar 
context for both answers and make 
appropriate assumptions when checking if 
they are the same. Reason step-by-step and 
put the final answerYES or NO in \boxed{}. 
\end{verbatim}
\\
\hline
\textbf{User Prompt:} \newline
\begin{verbatim}
Options:
A: <Option 1>
B: <Option 2>
C: <Option 3>
D: <Option 4>
answer 1: <Correct Answer>
answer 2: <Predicted Answer>
\end{verbatim}
&
\textbf{User Prompt:} \newline
\begin{verbatim}
answer 1: <Correct Answer>
answer 2: <Predicted Answer>
\end{verbatim}
\\
\hline
\end{tabular}
\caption{Prompts used for Answer Matching}
\label{tb: answer_matching}
\end{table*}
\begin{table*}[t]
\begin{tabular}{|p{\textwidth}|}
\hline
\begin{verbatim}
Clean and standardize math questions by removing multiple-choice options, normalizing 
the answer format, identifying dependencies, and determining the language. For any 
answers expressed in MathML, convert them to LaTeX. Conversion of MathML in the 
**question** is *not required* (but preserve LaTeX if already present). 
Additionally, provide a clear **step-by-step reasoning** explaining how each part of 
the output was derived.

### Instructions:

1. Identify and extract the core question text:
    * Remove all multiple-choice options (e.g., A–D or 1–4), ensuring the main question 
    remains grammatically and semantically intact.
    * Preserve existing LaTeX in the question.
    * Do **not** convert MathML in the question. It may be retained as-is.

2. Normalize the answer:
    * If the answer is given as an option label (e.g., "Answer: B"), replace it with the 
    corresponding value from the provided options.
    * If the answer is already a value, retain it.
    * If the answer is in MathML, convert it to LaTeX.

3. Determine dependency flags:
    * **Option-dependent:** Is the question understandable and solvable without access 
    to the answer options? Mark `True` if the question lacks key information without 
    them; otherwise, `False`.
    * **Diagram-dependent:** Does the question reference or rely on a diagram, figure, or 
    visual element? Mark `True` or `False`.

4. Identify the language:
    * Detect and report the language of the question text (e.g., `English`, `Hindi`, 
    `Tamil`, etc.).

5. Provide reasoning:
    * For each output field (question, answer, flags, language), include a clear 
    explanation of how the output was determined.
    * The reasoning should follow a logical step-by-step format, but does **not** need to 
    be wrapped in any special `<reason>` block.

# Output Format

<question> cleaned question </question>  
<answer> cleaned answer </answer>  
<option_dependent> True/False </option_dependent>  
<diagram_dependent> True/False </diagram_dependent>  
<language> detected language </language>

* All math in the **answer** must be in LaTeX.
* There should be **no references** to original option labels (e.g., "A", "1", or
"Option B").
* Ensure the cleaned question is coherent, self-contained, and grammatically correct.
* The reasoning can be in free-text form and must explain how each part of the output was 
derived.
\end{verbatim} \\
\hline
\end{tabular}
\caption{Prompt used for Question Cleaning (Part 1 of 2)}
\label{tb: question_cleaning}
\end{table*}

\begin{table*}[t]
\begin{tabular}{|p{\textwidth}|}
\hline
\begin{verbatim}
### Example 1
Input:
What is the derivative of \(x^2 + 3x + 5\)?  
A) \(2x + 3\)  
B) \(x + 3\)  
C) \(x^2 + 3\)  
D) \(2x + 5\)  
Answer: A  

Output:
<question> What is the derivative of \(x^2 + 3x + 5\)? </question>  
<answer> \(2x + 3\) </answer>  
<option_dependent> False </option_dependent>  
<diagram_dependent> False </diagram_dependent>  
<language> English </language>
\end{verbatim}

\begin{verbatim}
### Example 2
Input:
<p>Simplify the following expression:</p>
<math xmlns="http://www.w3.org/1998/Math/MathML">
  <mfrac>
    <msqrt>
      <msup><mi>a</mi><mn>2</mn></msup>
    </msqrt>
    <mi>a</mi>
  </mfrac>
</math>

<p>Options:</p>
1) <math xmlns="http://www.w3.org/1998/Math/MathML"><msqrt><mi>a</mi></msqrt></math>  
2) <math xmlns="http://www.w3.org/1998/Math/MathML"><mi>a</mi></math>  
3) <math xmlns="http://www.w3.org/1998/Math/MathML"><mfrac><mn>1</mn><mi>a</mi>
</mfrac></math>  
4) <math xmlns="http://www.w3.org/1998/Math/MathML"><mn>1</mn></math>

Answer: 1

Output:
<question> Simplify the following expression:
<math xmlns="http://www.w3.org/1998/Math/MathML">
  <mfrac>
    <msqrt>
      <msup><mi>a</mi><mn>2</mn></msup>
    </msqrt>
    <mi>a</mi>
  </mfrac>
</math>
</question>  
<answer> \sqrt{a} </answer>  
<option_dependent> False </option_dependent>  
<diagram_dependent> False </diagram_dependent>  
<language> English </language>
\end{verbatim} \\
\hline
\end{tabular}
\caption{Prompt used for Question Cleaning (Part 2 of 2)}
\end{table*}
\end{document}